\def\year{2022}\relax
\documentclass[letterpaper]{article} 
\usepackage{aaai22}  
\usepackage{times}  
\usepackage{helvet}  
\usepackage{courier}  
\usepackage[hyphens]{url}  
\usepackage{graphicx} 
\usepackage{natbib}  
\usepackage{caption} 
\DeclareCaptionStyle{ruled}{labelfont=normalfont,labelsep=colon,strut=off} 
\usepackage{algorithm}
\usepackage{algorithmic}

\usepackage{newfloat}
\usepackage{listings}
\floatstyle{ruled}
\newfloat{listing}{tb}{lst}{}
\floatname{listing}{Listing}
\nocopyright

\usepackage{graphicx}
\usepackage{verbatim}
\usepackage[dvipsnames]{xcolor}
\usepackage{subcaption}%
\usepackage{wrapfig}
\usepackage{listings}
\usepackage{enumitem}
\usepackage{tikz}
\usepackage[frozencache,cachedir=minted-cache]{minted}
\usepackage{hyperref}
\usepackage{booktabs}

\def\checkmark{\tikz\fill[scale=0.4](0,.35) -- (.25,0) -- (1,.7) -- (.25,.15) -- cycle;} 

\title{POGEMA: Partially Observable Grid Environment for Multiple Agents}
\author{
Alexey Skrynnik\textsuperscript{\rm 1}, Anton Andreychuk\textsuperscript{\rm 1}, Konstantin Yakovlev\textsuperscript{\rm 1}, Aleksandr Panov \textsuperscript{\rm 1}
}
\affiliations{
\textsuperscript{\rm 1}AIRI, Moscow, Russia \\
}

\usepackage{bibentry}

\newcommand{\pogema}{\textsc{pogema}}

\usepackage{xcolor}
\newboolean{showcomments}
\setboolean{showcomments}{true}

\ifthenelse{\boolean{showcomments}}
  {\newcommand{\nb}[3]{
  {\color{#2}\small\fbox{\bfseries\sffamily\scriptsize#1}}
  {\color{#2}\sffamily\small$\triangleright~$\textit{\small #3}$~\triangleleft$}
  }
  }
  {\newcommand{\nb}[3]{}
  }

\begin{document}

    \maketitle

    \begin{abstract}
        We introduce \pogema\footnote[1]{Code available at \href{https://github.com/AIRI-Institute/pogema}{https://github.com/AIRI-Institute/pogema}} a sandbox for challenging partially observable multi-agent pathfinding (PO-MAPF) problems . This is a grid-based environment that was specifically designed to be a flexible, tunable and scalable benchmark. It can be tailored to a variety of PO-MAPF, which can serve as an excellent testing ground for planning and learning methods, and their combination, which will allow us to move towards filling the gap between AI planning and learning.

    \end{abstract}

    \section{Introduction}

    Multi-agent pathfinding (MAPF) is a challenging problem with typical applications in video games, logistics, etc. The intrinsic assumption of what is called Classical MAPF~\cite{stern2019multi} is that there exists a central controller which possesses all the information about the agents and the environment. It is this controller that plans a set of collision free paths for all of the agents. Thus, such variant of MAPF can be deemed to be \emph{fully observable} and \emph{centralized}. Indeed, plenty of methods exist to solve this variant of MAPF, see~\cite{sharon2015conflict, cap2015a, surynek2009novel, Wagner2011} for example.

    In many practical applications, though, it is impossible to deploy a reliable infrastructure for a centralized MAPF. Consider, for example, a  coverage/surveillance of areas such as nuclear plants, mines, disaster areas, etc. These areas can have poor communications (or none at all). Thus, they require a fundamentally different variant of MAPF, i.e. on the one when the central controller is absent and each agent has limited communication/observation capabilities -- \emph{partially observable} MAPF (PO-MAPF), which is intrinsically \emph{decentralized}.

\begin{figure}[t]
        \centering

        \includegraphics[width=1.0\columnwidth]{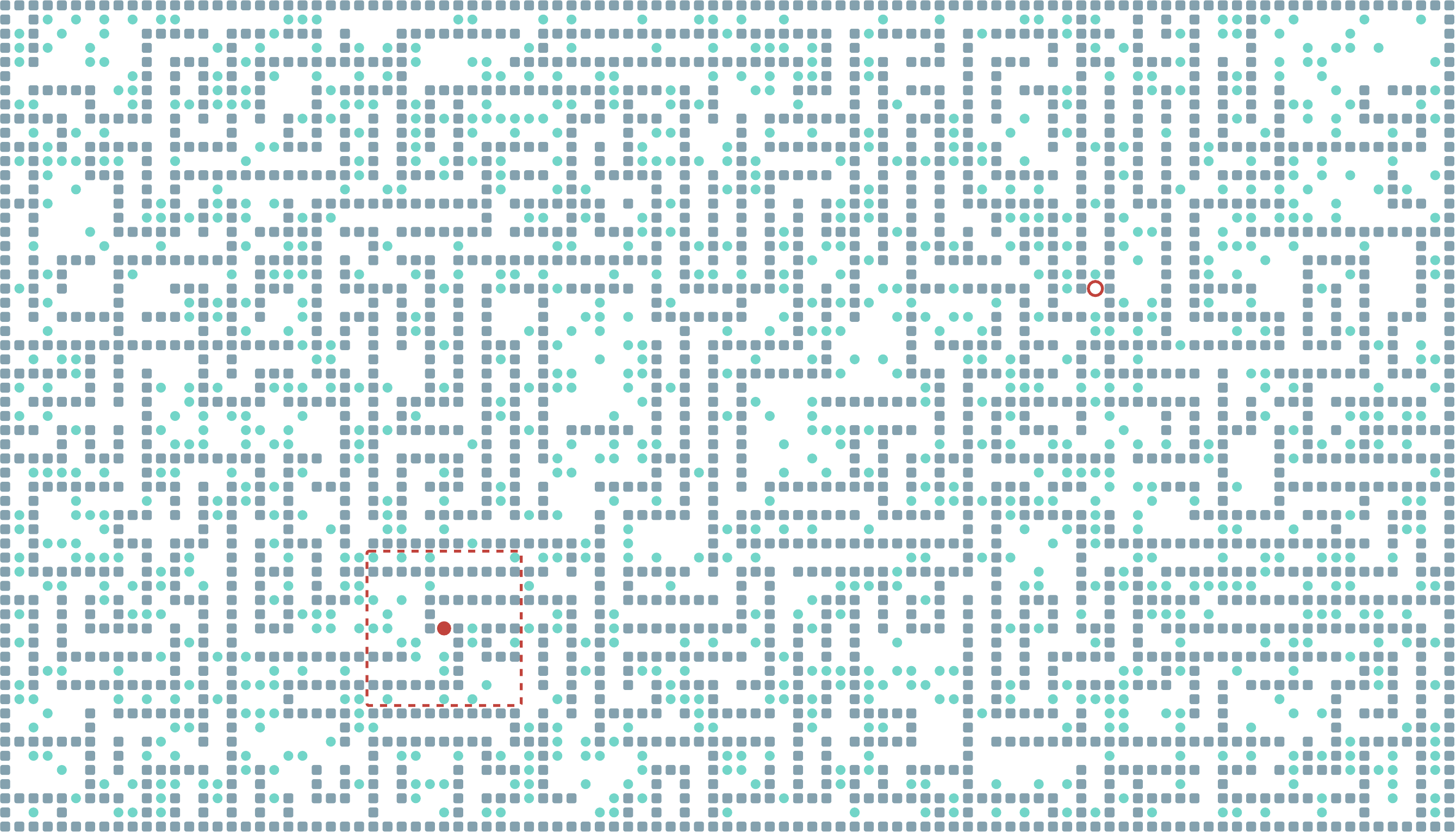}

        \caption{A typical PO-MAPF instance in \pogema. The agent for whom the path is being constructed is indicated by a red circle, his goal is a white circle with a red border, his area of observation is a square with an intermittent red border. Static obstacles are shown in gray,  other agents -- in turquoise.}
        \label{fig:pogema}
    \end{figure}

    Indeed, PO-MAPF requires different methods compared to fully observable MAPF. In the former case, we seek not for a fixed solution, i.e. the set of conflict-free plans, but rather for a policy that maps agents' observations to actions in such a way that it maximizes the chance of reaching the goal while avoiding the collisions and minimizing the number of actions performed. To foster the development of such methods, which are likely to span across different areas of AI (Search, Planning, Learning etc.) a fast and flexible software environment is needed, which will allow researchers to quickly prototype and evaluate the methods for solving PO-MAPF problems. In this work we present such an environment -- \pogema~(Partially Observable Grid Environment for Multiple Agents) (see an example on figure~\ref{fig:pogema}).
    
    \pogema~relies on the widespread grid representation of the surrounding space and objects. As dictated by PO-MAPF, each agent in this environment has an access only to a local observation, i.e. ego-centric patch of the grid of the user-defined size. Relying on the (history of) observations an agent chooses an action to be performed at the next time step. \pogema~is equipped with several baseline policies for choosing such an action, including the search-based one and the learning-based one. Indeed, \pogema~allows plugging-in user-designed policies both for evaluation and for training (in case of the learnable policies).

    \pogema~is specifically tailored to train  policies based on Reinforcement Learning (RL) methods. It provides interfaces for  well-known RL frameworks and its sample efficiency is up to 10k FPS (frame per second) even for single-agent cases (for multi-agent cases it is much faster). The latter allows fast training of the complex value and state approximators based on neural network models.

    Some preliminary results show that some heuristic search-based planners are sufficient to solve some PO-MAPF tasks, and learnable approaches are better able to cope with other types of tasks. This shows that PO-MAPF setting can serve as an excellent testing ground for the development of hybrid methods of simultaneous planning and learning, which will allow us to move towards filling the gap between AI Planning and Reinforcement Learning. \pogema~is an excellent tool for testing the efficiency of new methods in this area and presents a fast and effective tool and a set of baselines with comprehensible metrics and a simple interface for creating your own new algorithms.

    \section{Related Works}
    There are a number of environments in the RL community that are used to train agents in a multi-agent setting (MARL). Such environments include Flatland~\cite{mohanty2020flatland} and Petting Zoo MAgent~\cite{magent}. Flatland is designed to solve the specific problem of fast conflict-free train scheduling on a fixed railway map. This environment is quite slow and focused on full observability. MAgent is a fast environment for modelling the group and swarm behaviour of agents with a set of actions that, in addition to moving, includes actions for interacting with other agents. This environment has a limited set of scenarios and does not have an interface for testing planning solutions. A number of other visual-based  environments known in MARL (Dota 2, Starcraft) seem to be more heavy and complex in terms of encoding observations and actions, which makes it impossible to give a quick and convenient interface for comparing trajectory planning methods. Table~\ref{table:compare} provides a brief comparison of the main features of \pogema~and similar environments.

    \begin{table}[htb!]
        \centering
        \caption{Comparison of \pogema~ with other multi-agent grid-based environments. FPS was benchmarked with 80 agents for each environment. For MAgent we used the \textit{Battlefield} task. }
        \label{table:compare}
        \resizebox{\linewidth}{!}{
        \begin{tabular}{lrccc}
                        &     & Procedural  & Requires       & Partial \\
            Environment & FPS & generation  & generalization & observability \\
            \midrule
            \pogema & 83.000  & \checkmark  & \checkmark  & \checkmark \\
            MAgent & 184.000 & &  & \checkmark \\
            Flatland & 156 & \checkmark & \checkmark & \\
        \end{tabular}
        }
    \end{table}

    \section{\pogema~Environment}
    
    \subsection{Basics}
    Consider $n$ agents which populate to the 4-connected grid composed of the free and blocked cells. This grid can be either procedurally generated by \pogema~or provided by the user. Each agent is assigned to the goal cell which it has to reach. At every time step an agent receives a local ego-centric observation, whose size $R$ is defined by the user, and picks an action, which can be either move to one of the adjacent cells or waits in the current cell. The action picking algorithm, i.e. the \textit{policy}, is defined by the user (several baselines are also provided). After each agent picks an action \pogema~applies all the actions that are feasible, i.e. do not lead to the collisions (either between the agent and the obstacle or between several agents). Agents that picked the infeasible action remain where they were. If an agent enters its goal cell it is removed from the environment (the so-called ``disappear-at-target'' behavior). The \textit{episode} ends either if all the agents reach their goals or if the user-specified time step limit, $K$, is reached.

    \subsection{Observation space}
        
        In \pogema, the agent occupying the cell with the coordinates $(i, j)$ is able to observe the status of the cells $i \pm R, j \pm R$, where $R$ is the user-defined observation radius. 
        Thus, the observation is a patch of a grid of size $[2\cdot R+1] \times [2 \cdot R+1]$ centred at the currently occupied cell. 
        Any information regarding the other agents, except their current locations (e.g., their goals, paths (or path segments) to the goals, etc.), \emph{is not included} in the observation. This is purposefully done to simulate the most challenging PO-MAPF setup, when the agents can not communicate with each other and share any information. We are planning to add other, less restrictive, observation designs in future.

        \begin{figure}[ht!]
            \centering
    
            \includegraphics[width=\columnwidth]{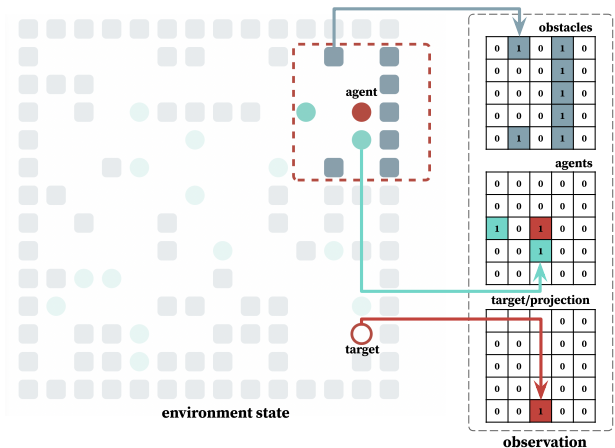}
            \caption{\pogema observation space for RL algorithms. Observation space consists of three matrices corresponding to obstacles, other agents and target or its projection.}
            \label{fig:observation}
        \end{figure}
        
    Technically, the observation is encoded as three matrices. The first one bears the information about the static obstacles within the field of view. The second one contains the information about the other agents. The third one includes the goal projection (see Figure~\ref{fig:observation}). The agent's target can be encoded using direction or relative coordinates, but we present it as a matrix to simplify input encoding of the neural network (it is easier to encode data of one modality).

    \subsection{State Space}
    Majority of MARL algorithms utilise environment state during training (centralized training, decentralized execution scheme). Pogema has a method to provide state of the environment. For PO-MAPF the full state consists of the global map and the positions/goals of all agents. In contrast to observation space, the state space can change depending on the number of agents and size of the grid. Thus, algorithms with centralized training are restricted to be trained in the exact PO-MAPF domain.

    \subsection{Reward Function and Metrics}
    
    The agent receives a reward of $1.0 $ when it reaches the goal and $ 0.0 $ in all other cases.
    We have chosen this function, since it is universal for the wide range of PO-MAPF configurations. Moreover, this reward is easy to interpret, as it corresponds to the individual success rate (ISR) metric for each agent. To shape the reward function one could use the wrappers mechanism of OpenAI Gym. The second metric is a cooperative success rate (CSR), which is equal to $1$ only if all of the agents have reached the target.

    \subsection{Interfaces for RL Frameworks}

    \textit{OpenAI Gym}~\cite{brockman2016openai}  is de facto standard interface for the agent to interact with the environment in RL. Unfortunately, Gym is designed for classic single-agent interaction, which restricts its application for PO-MAPF problems. \textit{PettingZoo}~\cite{terry2020pettingzoo} framework provides a unified interaction interface for multi-agent RL (MARL).

    Pogema provides PettingZoo and single-agent Gym integration out of the box (see example Figure~\ref{code:petting-zoo}).
    Besides PettingZoo, MARL community has several other interfaces: vectorized Gym-like interaction, interaction used in PyMARL~\cite{samvelyan19smac} and Rllib~\cite{rllib} interface. In addition to these frameworks we provide integration with high-quality Asynchronous-PPO algorithm, implemented in \textit{SampleFactory}~\cite{petrenko2020sample}.

    \begin{figure}[htb!]
        \footnotesize
\begin{minted}{python}
import gym
import pogema

# Create Pogema environment
# with PettingZoo interface
env = gym.make("Pogema-8x8-hard-v0",
integration="PettingZoo")
...

# Create single-agent Pogema environment
# with OpenAI Gym interface
env = gym.make("Pogema-8x8-easy-v0",
integration="gym")
\end{minted}
        \caption{Sample code for creating Pogema environment with a parallel PettingZoo interface and single-agent Gym interface. Please note, \textit{Pogema-8x8-easy-v0} corresponds to a grid with size $8\times8$ with only one agent.
        \label{code:petting-zoo}}
    \end{figure}

    \subsection{Builtin Benchmarks}
    \begin{figure}[htb!]
        \centering
        
        \begin{subfigure}[b]{0.45\columnwidth}
            \includegraphics[width=\textwidth]{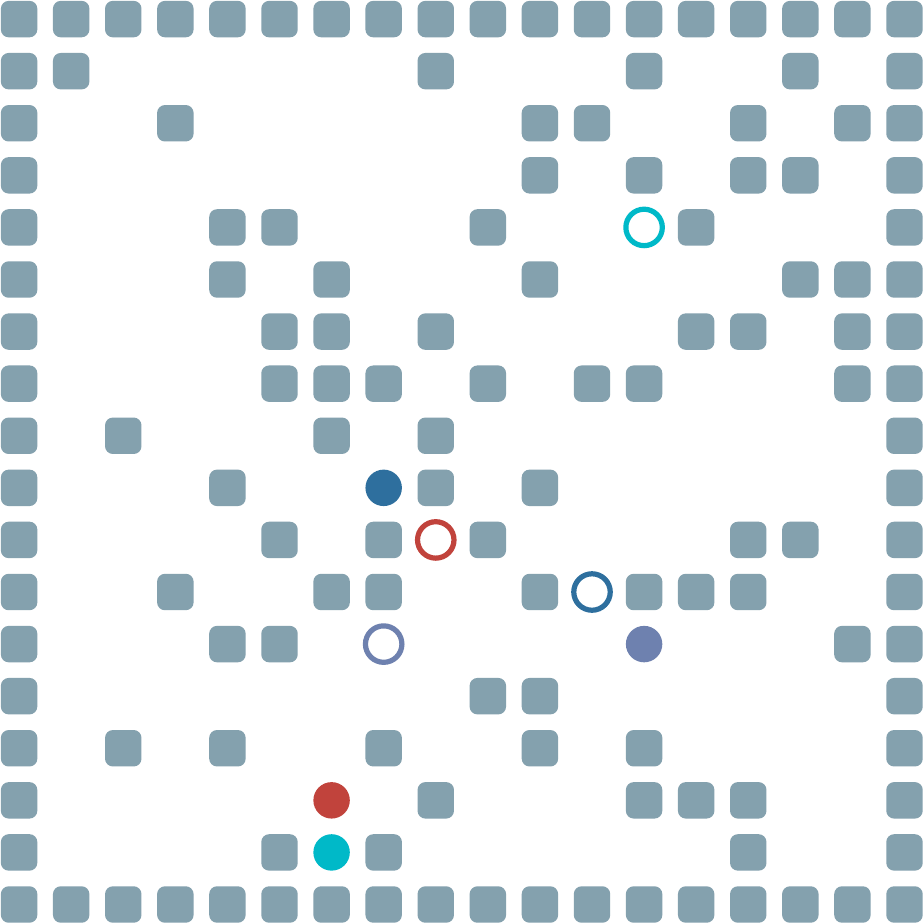}
            \caption*{Pogema-16x16-easy-v0}
        \end{subfigure}
        \hspace{0.03\textwidth}
        \begin{subfigure}[b]{0.45\columnwidth}
            \includegraphics[width=\textwidth]{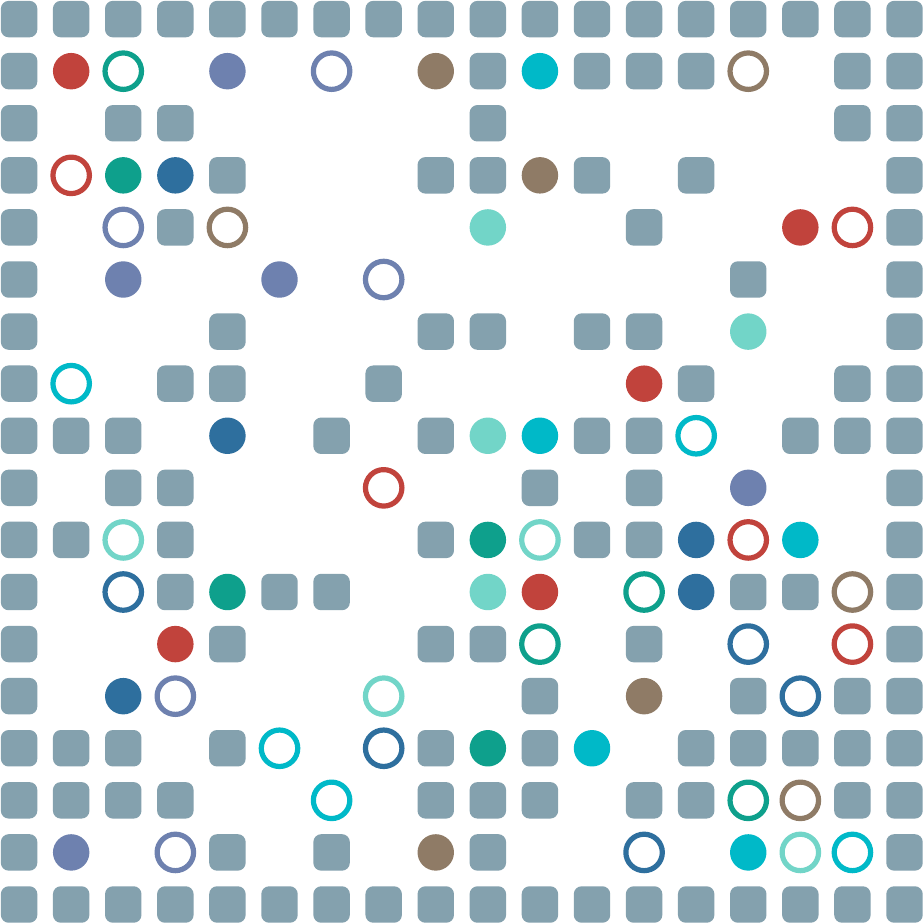}
            \caption*{Pogema-16x16-extra-hard-v0}
        \end{subfigure}
        \caption{Easy and extra-hard Pogema configuration for $16\times16$ map. }
        \label{fig:difficulty}
    \end{figure}
    
    We provide builtin configurations of the maps for easier comparison of different algorithms.
    Pogema benchmark consists of three scales of grid configurations with four levels of difficulty (see example in Figure~\ref{fig:difficulty}). We picked sizes: $8\times8$, $16\times16$, $32\times32$, $64\times64$. The obstacles are placed randomly with density $30\%$.
    Maps with obstacle density between $30\%$ and $40\%$ present the greatest difficulty, since maps with less obstacles are trivial and maps with more obstacles are decomposed to several components. It's guaranteed that any agent can solve the task, if other agents do not interfere him. We provide four difficulty levels: \textit{easy}, \textit{normal}, \textit{hard} and \textit{extra-hard}, which corresponds to different density of agents on the map (see Table~\ref{table:configs}). For all these environments agent's the observation  radius is the same, and equals to $5$, which corresponds to $11x11$ field of view.

    \subsection{Custom Maps}

    \pogema~provides a wide range of customization options. 
    First, we describe how to adjust random maps.  
    The main settings are defined in \textit{GridConfig}: size of the environment, obstacle density, number of agents, radius of agent’s field of view and maximum length of the episode. In addition \textit{GridConfig} has seeding option, which is \textit{None} by default (thus, positions of obstacles, agents and targets are new after each reset).  
    
        \begin{figure}[htb!]

        \footnotesize
\begin{minted}{python}
import gym
from pogema import GridConfig

# Define four rooms map as a string
grid = """
.....#.....
.....#.....
...........
.....#.....
.....#.....
#.####.....
.....###.##
.....#.....
.....#.....
...........
.....#.....
"""

# Define new configuration with 8 randomly placed agents
grid_config = GridConfig(map=grid, num_agents=8)

# Create Pogema environment using Gym
env = gym.make('Pogema-v0', grid_config=config)
\end{minted}
        \caption{A sample code for creating PO-MAPF instance with custom map in Pogema.
        \label{code:custom-map}}
    \end{figure}
    
    To create an environment with a custom map, one is supposed to specify \textit{map} field in \textit{GridConfig} (see Figure~\ref{code:custom-map}). Positions of agents and targets also can be defined in \textit{GridConfig} if that needed. For the large configurations we suggest the option to store it in \textit{YAML} format, which will be validated on \pogema~ side.

    \section{Experiments}
    
    \subsection{Planning Baseline}
    
    PO-MAPF problems can be solved with varied success by classical path-planning approaches such as A*\cite{hart1968}. Each agent plans its path independently, replanning the path on each iteration, receiving new observation. Other agents are considered as static obstacles, as their trajectories are unknown and they cannot communicate. 
    
    While pure A* can easily find a path for a single agent, in case of presence of large amount of agents on the map, they block each others paths. To increase the chances to find a solution, the suggested planning baseline uses some ad-hoc enhancements. First, in case of failure to find a trajectory for an agent, it moves to the adjacent cell that is the closest to the goal. Second, in case of indicating an oscillating behavior, when agent repeatedly moves between the same cells, a wait action with 50\% probability is added.
    
    The results of the planning baseline with different combinations of enhancements are presented in Table \ref{table:replan}.
    
    \subsection{PyMARL Baselines}
\begin{figure}[ht!]
        \centering
       \includegraphics[width=\columnwidth]{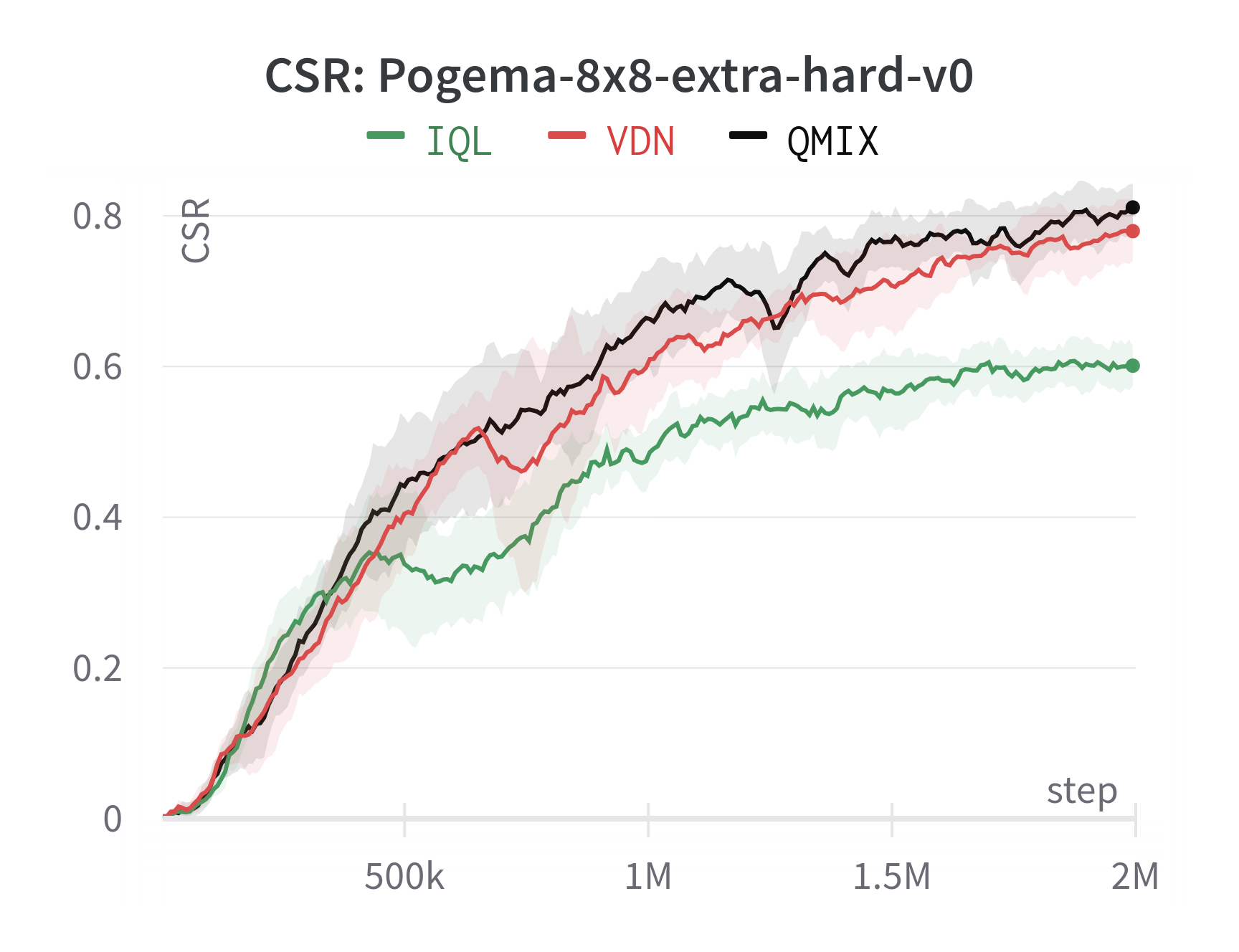}
       
        \caption{Learning curves of PyMARL algorithms. Centralized methods (QMIX, VDN) show better results, since \textit{Extra-hard} version of \pogema~$8\times8$ requires cooperative interaction. The results are averaged over three runs.}
        \label{fig:pymarl}
    \end{figure}

    Centralized methods are the standard techniques in MARL tasks. In this experiment we show \pogema capabilities for benchmarking such kinds of algorithms.
    We use PyMARL implementations of such algorithms as: QMIX~\cite{rashid2018qmix}, VDN~\cite{sunehag2017value} (see Figure~\ref{fig:pymarl}). Also, we provide results for decentralized approach IQL~\cite{tan1993multi}.  
    For the experiment we used \textit{extra-hard} version of $8\times8$ environment and trained each algorithm for 2 million steps. The results emphasize the requirement of cooperative behaviour of the agents.

    \subsection{APPO Baseline}

\begin{figure}[htb!]
        
        \centering
            
        \begin{subfigure}[b]{0.45\textwidth}
            \includegraphics[width=\textwidth]{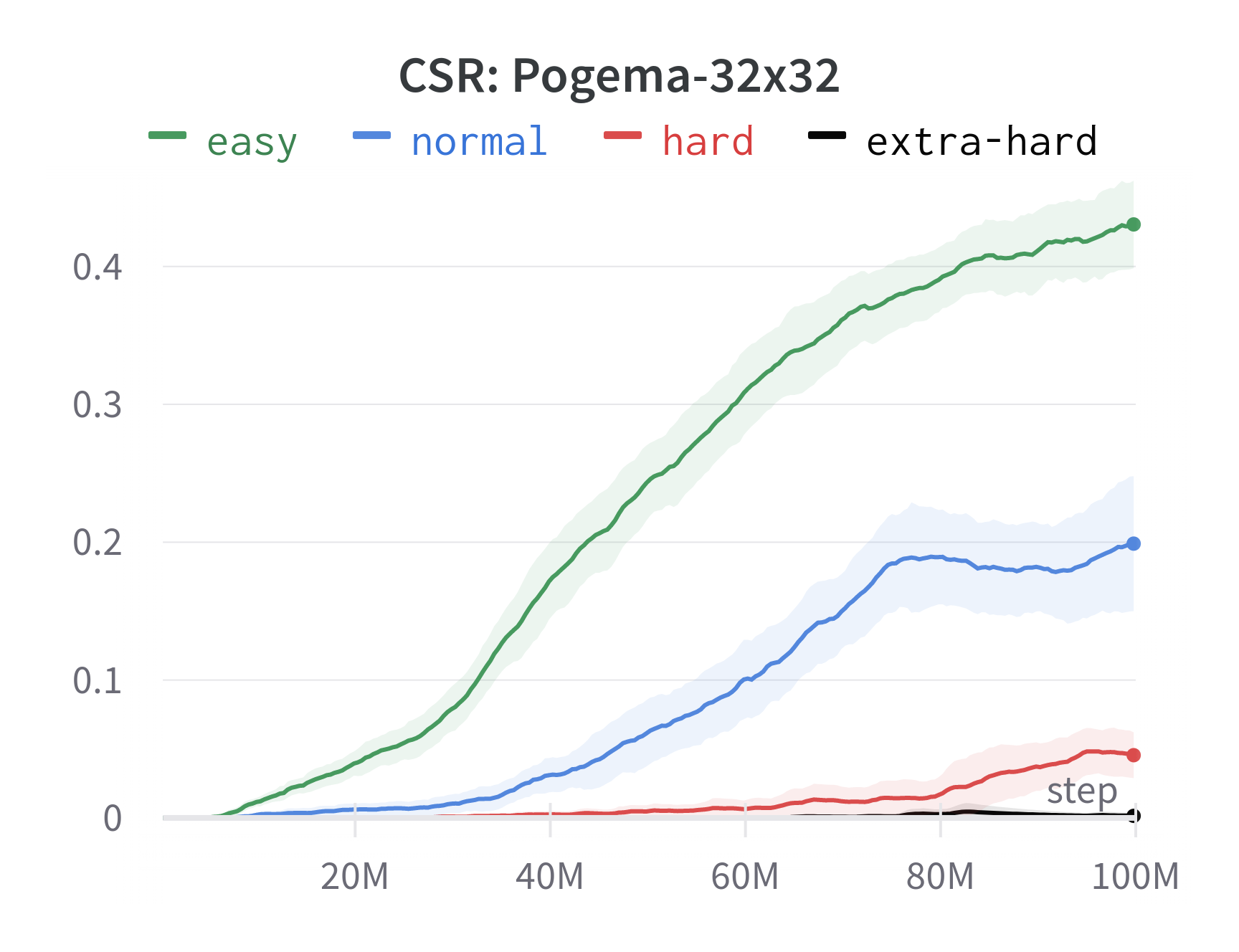}
        \end{subfigure}
        \caption{CSR metrics for APPO trained on $32\times32$ builtin \pogema~ configurations.}
        \label{fig:appo}
        
    \end{figure}

    Besides small PO-MAPF tasks, \pogema is suitable for large-scale RL experiments. Figure~\ref{fig:appo} presents training results for all difficulties of $32x32$ configurations, trained with 100 millions steps. For this experiment, we use \textit{SampleFactory} implementation of APPO algorithm. Fast environment and distributed RL framework allows us to train RL agent in a few hours on a single GPU, even with Impala~\cite{espeholt2018impala} encoder and recurrent heads. 
    Moreover, in $32x32$ \textit{extra-hard} configuration were simultaneously trained $128$ agents in each environment worker. Increasing  density of the agents and map size, significantly decreases the performance of the algorithms. Which shows the importance of both   pathfinding and conflict resolution components.

    \section{Conclusion}
    
    In this work we presented \pogema~ fast and easy to use environment for creating  a variety of PO-MAPF tasks.   We've designed a number of builtin configurations to help the community benchmark both learning and planning approaches. \pogema~ environments are procedurally generated, which ensures agent's ability to generalization. To simplify further experimentation we open source our code\footnote{Code available at \href{https://github.com/Tviskaron/pogema-baselines}{github.com/Tviskaron/pogema-baselines}} with RL and planning algorithms.

\bibliography{aaai22}

\appendix
\section{Appendix}

\begin{table}[htb!]
        \centering
        \caption{Builtin \pogema~ configurations.}
        \label{table:configs}
        \resizebox{\linewidth}{!}{
        \begin{tabular}{lccc}
            & agent  & num    & Episode          \\
            Environment                         & density & Agents & length \\
            \midrule
            \textit{Pogema-8x8-easy-v0}         & 2.2\%   & 1      & 64      \\
            \textit{Pogema-8x8-normal-v0}       & 4.5\%   & 2      & 64      \\
            \textit{Pogema-8x8-hard-v0}         & 8.9\%   & 4      & 64      \\
            \textit{Pogema-8x8-extra-hard-v0}   & 17.8\%  & 8      & 64      \\
            \midrule
            \textit{Pogema-16x16-easy-v0}       & 2.2\%   & 4      & 128     \\
            \textit{Pogema-16x16-normal-v0}     & 4.5\%   & 8      & 128     \\
            \textit{Pogema-16x16-hard-v0}       & 8.9\%   & 16     & 128     \\
            \textit{Pogema-16x16-extra-hard-v0} & 17.8\%  & 32     & 128     \\
            \midrule
            \textit{Pogema-32x32-easy-v0}       & 2.2\%   & 16     & 256     \\
            \textit{Pogema-32x32-normal-v0}     & 4.5\%   & 32     & 256     \\
            \textit{Pogema-32x32-hard-v0}       & 8.9\%   & 64     & 256     \\
            \textit{Pogema-32x32-extra-hard-v0} & 17.8\%  & 128    & 256     \\
            \midrule
            \textit{Pogema-64x64-easy-v0}       & 2.2\%   & 64     & 512     \\
            \textit{Pogema-64x64-normal-v0}     & 4.5\%   & 128    & 512     \\
            \textit{Pogema-64x64-hard-v0}       & 8.9\%   & 256    & 512     \\
            \textit{Pogema-64x64-extra-hard-v0} & 17.8\%  & 512    & 512     \\

        \end{tabular}
        }
    \end{table}
    
\begin{table}[htb!]
        \centering
        \caption{Percentage of successfully solved instances by planning baseline. GA - greedy actions. FL - fix loops.}
        \label{table:replan}
        \resizebox{\linewidth}{!}{
        \begin{tabular}{lcccc}
            & A*  & A*+GA    & A*+FL & A*+GA+FL          \\
            \midrule
            \textit{Pogema-8x8-easy-v0}         &100\%	&	100\%	&	100\%	&	100\% \\
            \textit{Pogema-8x8-normal-v0}       &100\%	&	100\%	&	100\%	&	100\% \\
            \textit{Pogema-8x8-hard-v0}         & 82\%	&	84\%	&	90\%	&	100\% \\
            \textit{Pogema-8x8-extra-hard-v0}   & 60\%	&	64\%	&	66\%	&	92\% \\
            \midrule
            \textit{Pogema-16x16-easy-v0}       &96\%	&	96\%	&	98\%	&	100\% \\
            \textit{Pogema-16x16-normal-v0}     &68\%	&	78\%	&	96\%	&	100\% \\
            \textit{Pogema-16x16-hard-v0}       &46\%	&	50\%	&	86\%	&	100\% \\
            \textit{Pogema-16x16-extra-hard-v0} & 10\%	&	14\%	&	38\%	&	84\%\\
            \midrule
            \textit{Pogema-32x32-easy-v0}       & 38\%	&	38\%	&	98\%	&	98\%\\
            \textit{Pogema-32x32-normal-v0}     & 12\%	&	16\%	&	94\%	&	96\%\\
            \textit{Pogema-32x32-hard-v0}       & 0\%	&	0\%	&	62\%	&	80\%\\
            \textit{Pogema-32x32-extra-hard-v0} & 2\%	&	2\%	&	6\%	&	22\%\\
        \end{tabular}
        }
\end{table}

\begin{figure*}[ht!]
        \centering
        
        \begin{subfigure}[b]{0.31\textwidth}
            \includegraphics[width=\textwidth]{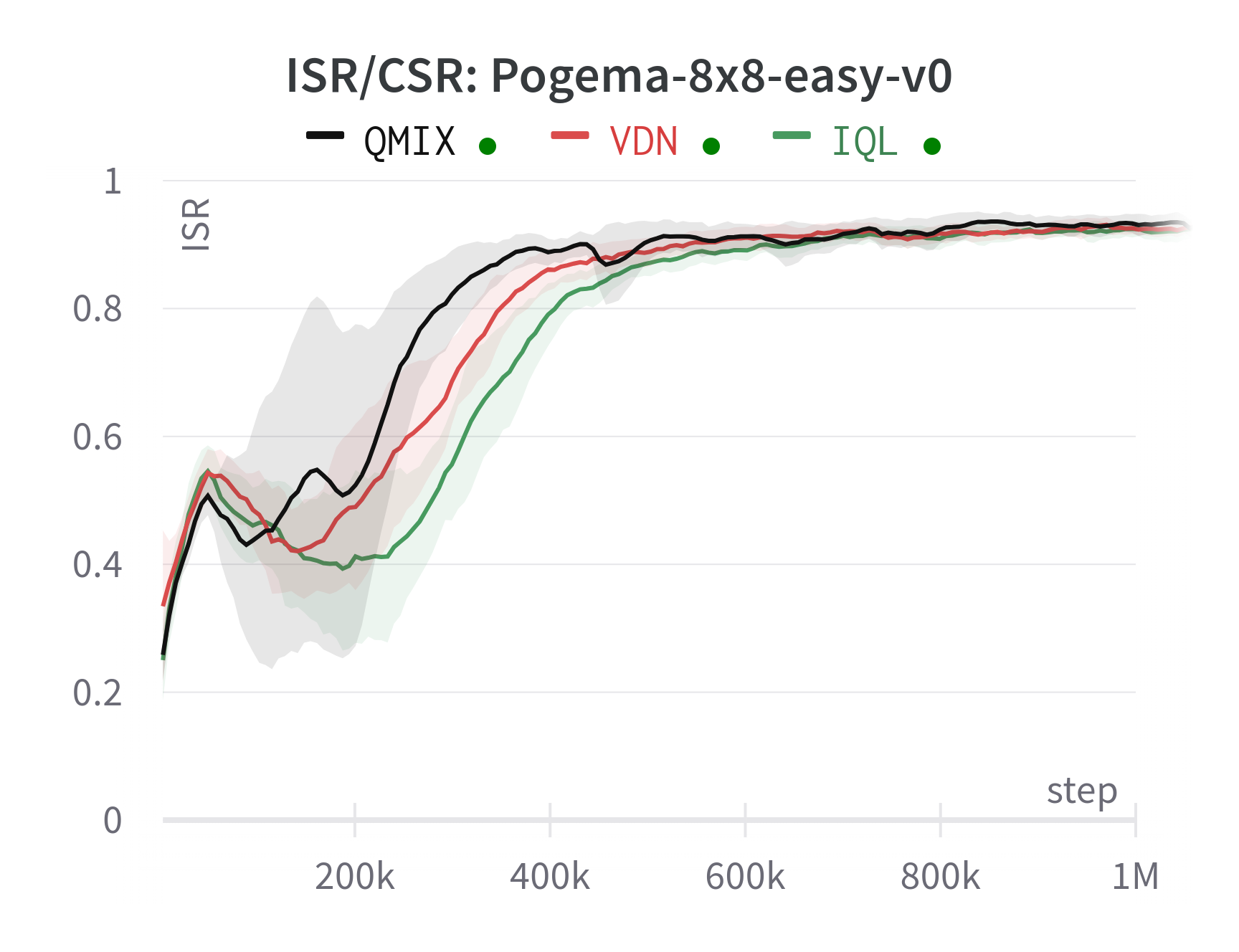}
        \end{subfigure}
        \hspace{0.03\textwidth}
        \begin{subfigure}[b]{0.31\textwidth}
            \includegraphics[width=\textwidth]{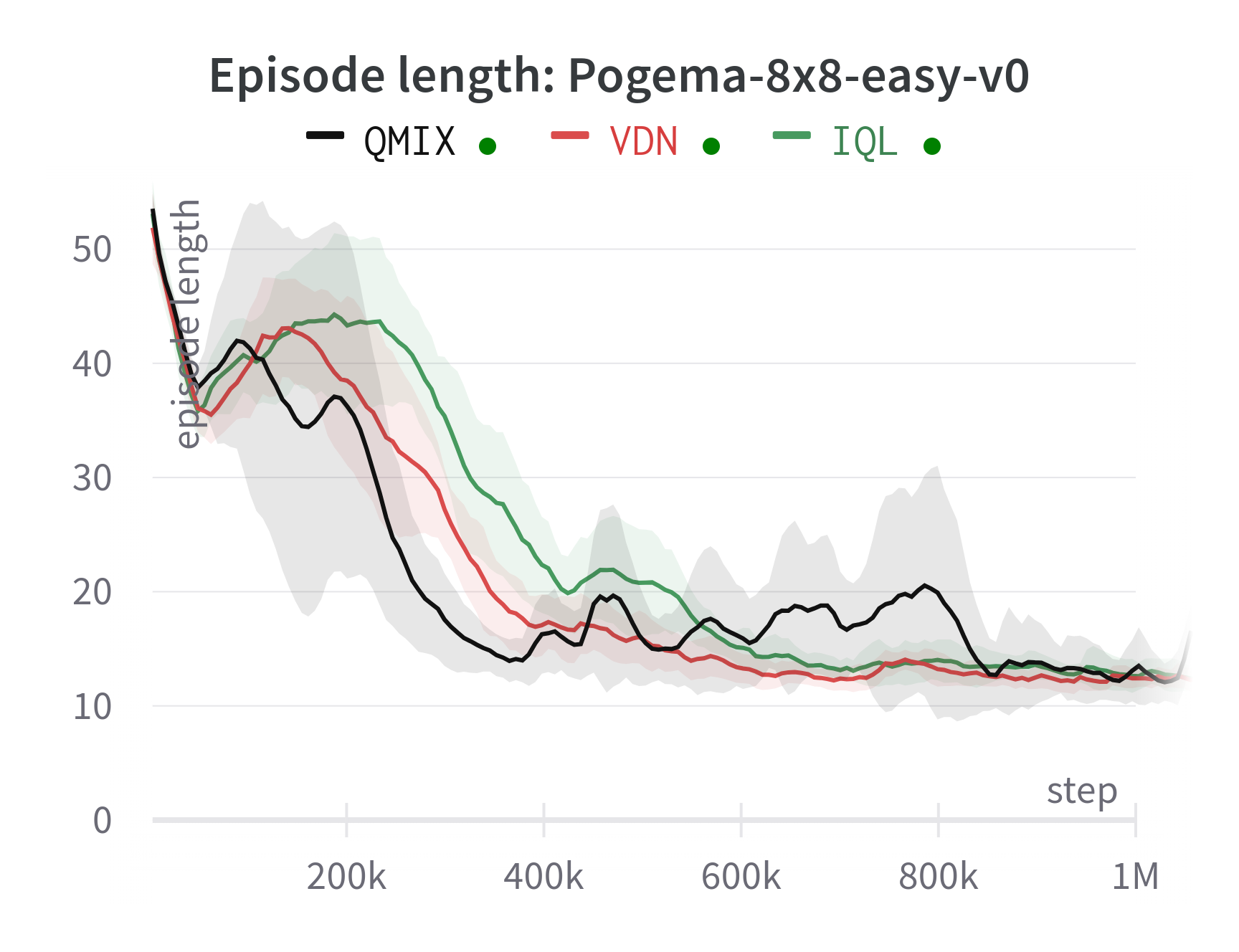}
        \end{subfigure}
        
        \begin{subfigure}[b]{0.31\textwidth}
            \includegraphics[width=\textwidth]{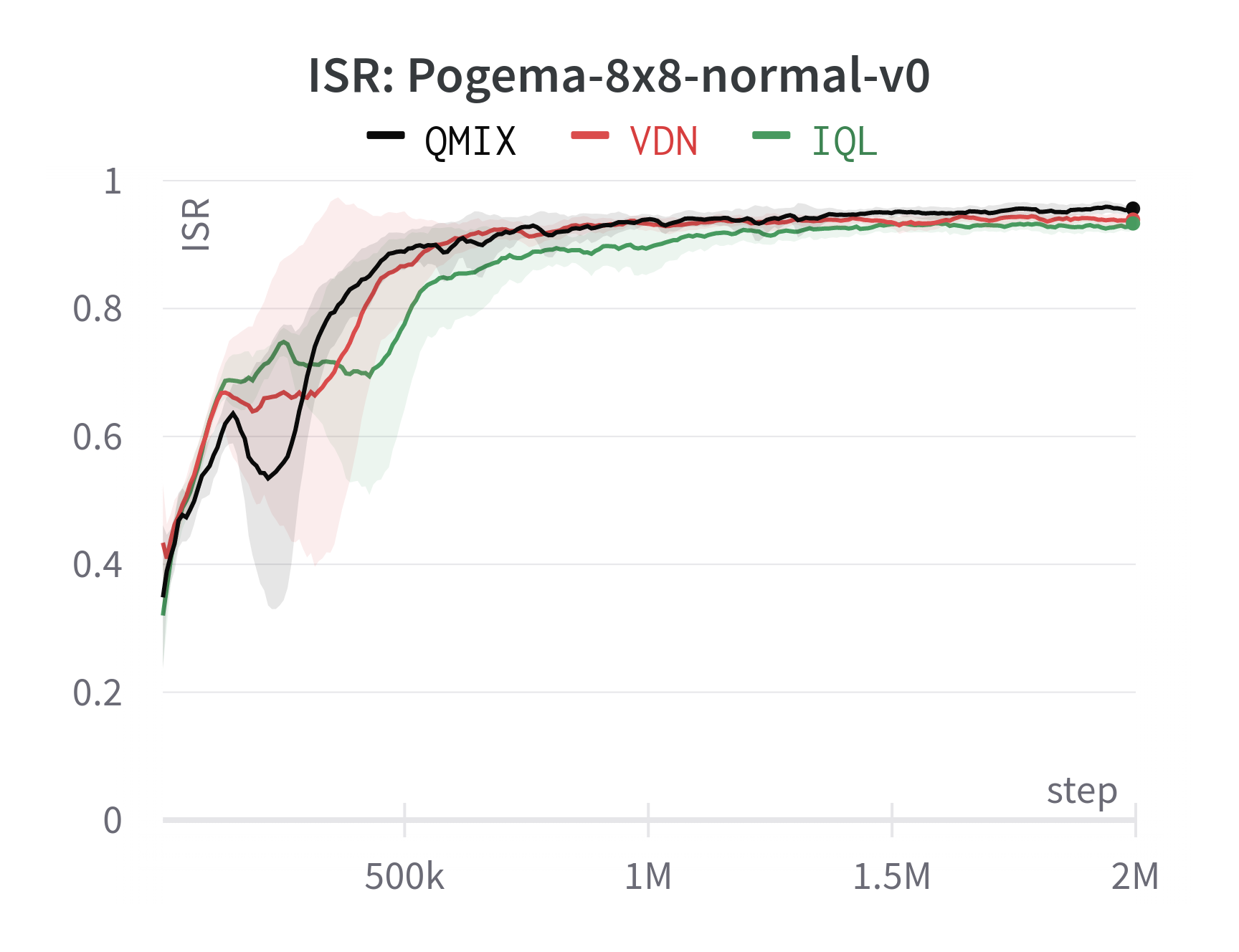}
        \end{subfigure}
        \hspace{0.01\textwidth}
        \begin{subfigure}[b]{0.31\textwidth}
            \includegraphics[width=\textwidth]{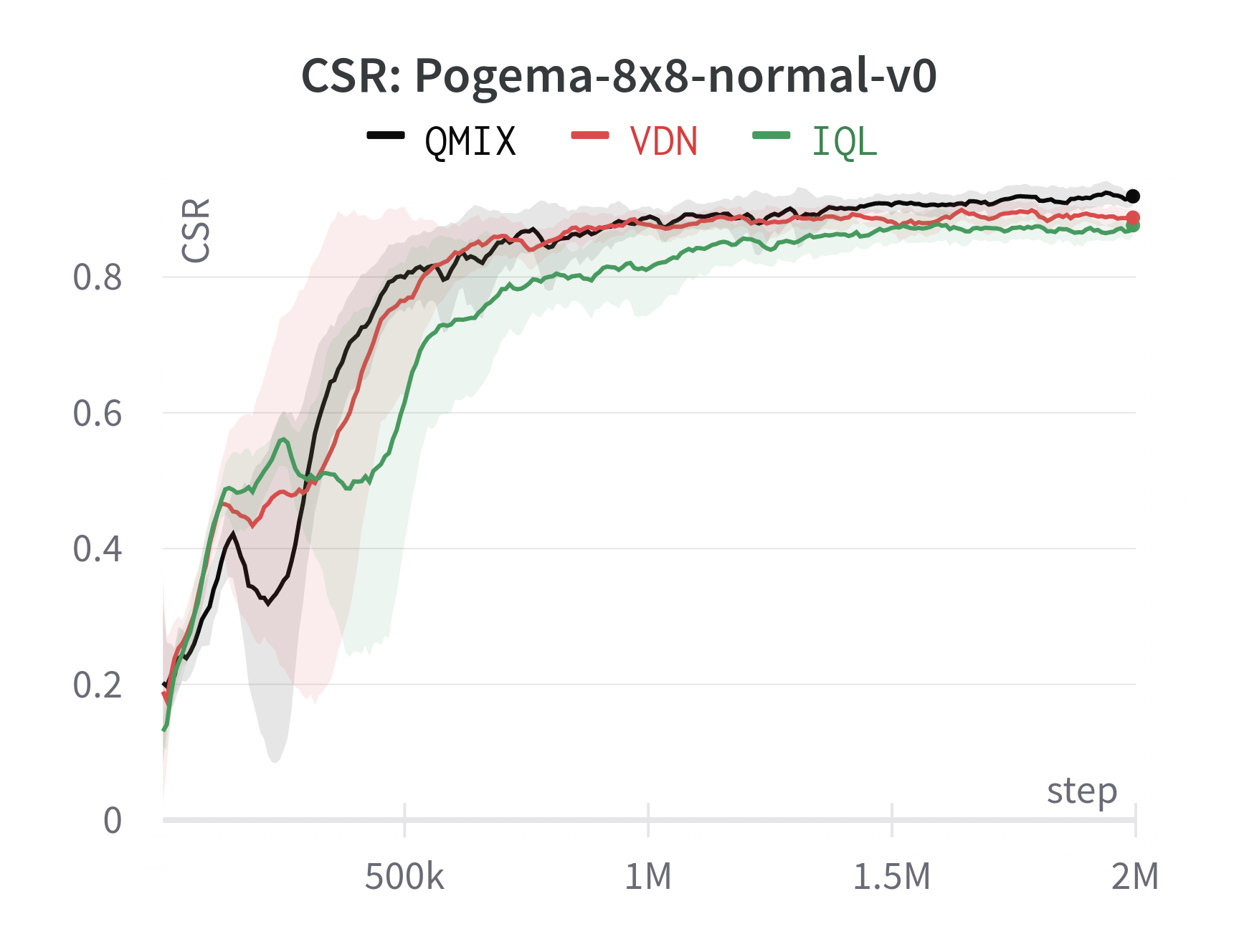}
        \end{subfigure}
        \hspace{0.01\textwidth}
        \begin{subfigure}[b]{0.31\textwidth}
            \includegraphics[width=\textwidth]{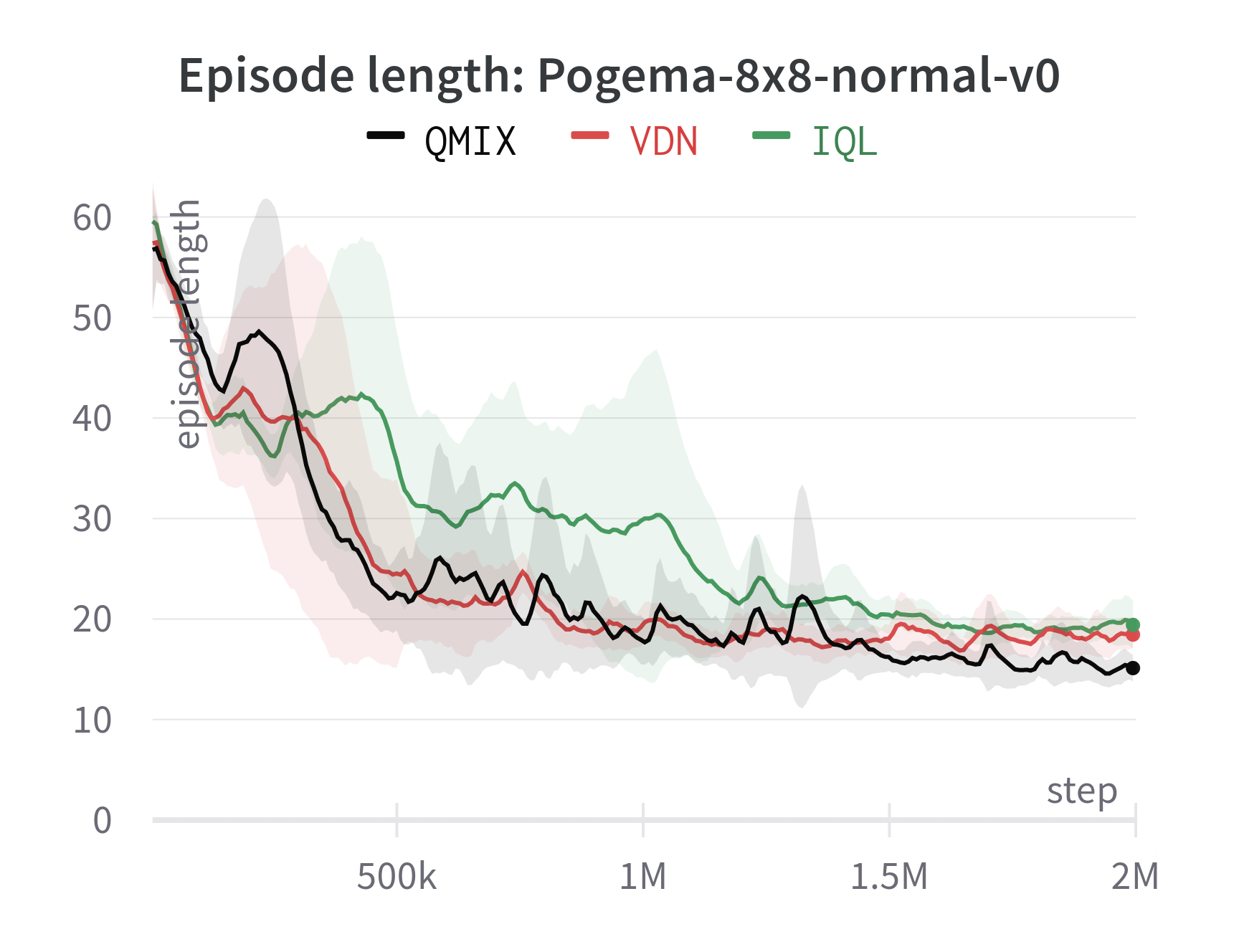}
        \end{subfigure}
        
        \begin{subfigure}[b]{0.31\textwidth}
            \includegraphics[width=\textwidth]{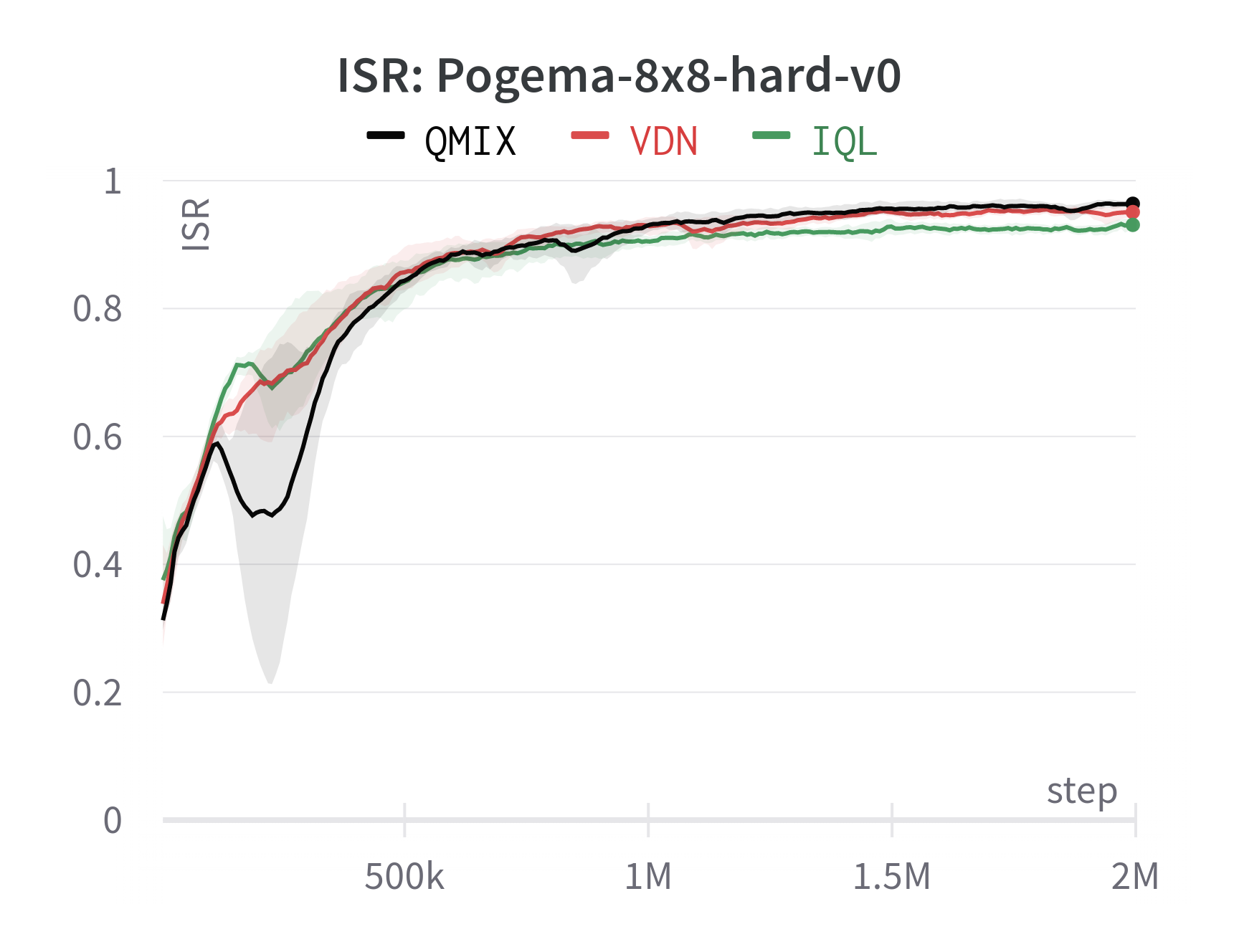}
        \end{subfigure}
        \hspace{0.01\textwidth}
        \begin{subfigure}[b]{0.31\textwidth}
            \includegraphics[width=\textwidth]{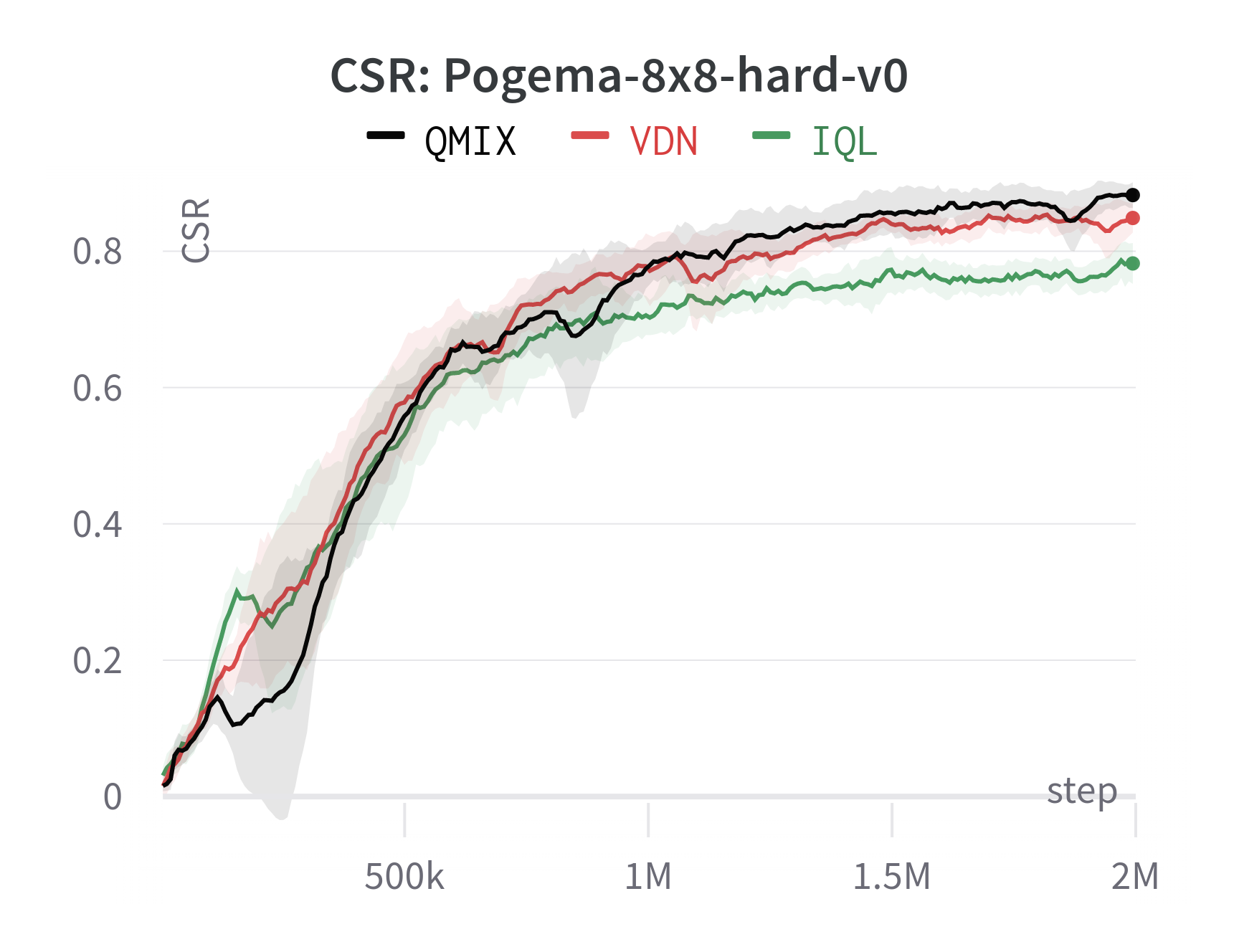}
        \end{subfigure}
        \hspace{0.01\textwidth}
        \begin{subfigure}[b]{0.31\textwidth}
            \includegraphics[width=\textwidth]{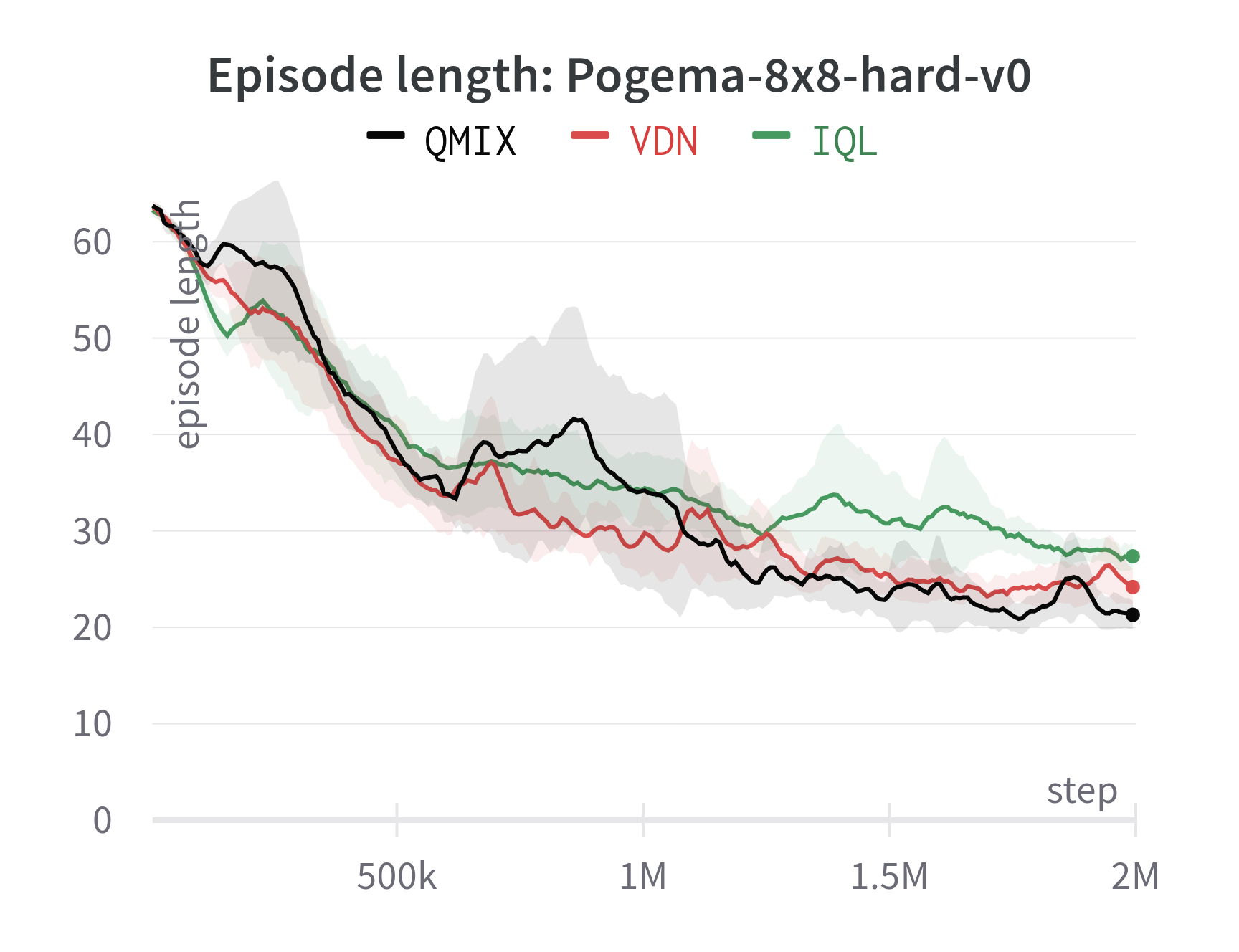}
        \end{subfigure}
        
        \begin{subfigure}[b]{0.31\textwidth}
            \includegraphics[width=\textwidth]{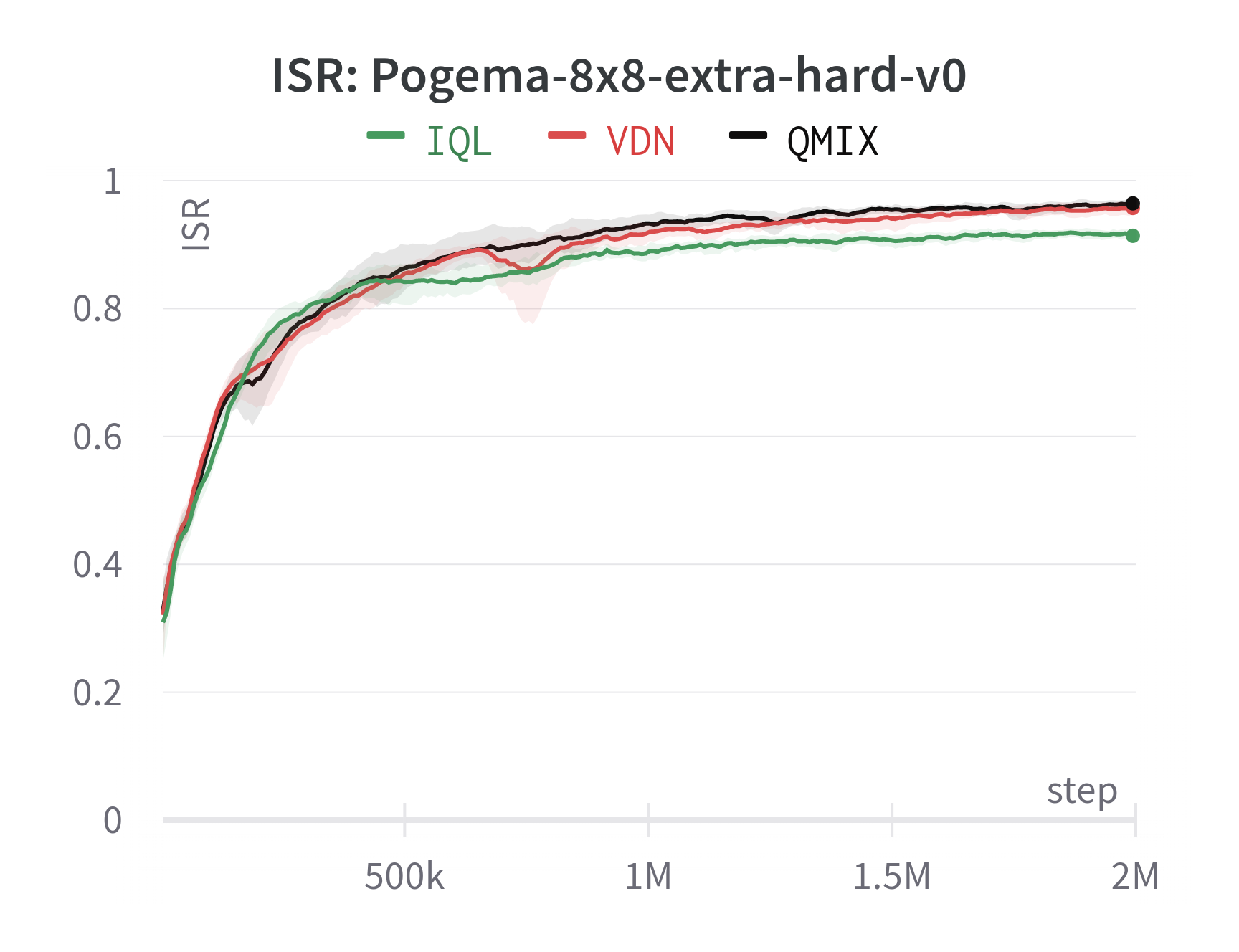}
        \end{subfigure}
        \hspace{0.01\textwidth}
        \begin{subfigure}[b]{0.31\textwidth}
            \includegraphics[width=\textwidth]{figures/pymarl/pymarl-extra-hard-csr.png}
        \end{subfigure}
        \hspace{0.01\textwidth}
        \begin{subfigure}[b]{0.31\textwidth}
            \includegraphics[width=\textwidth]{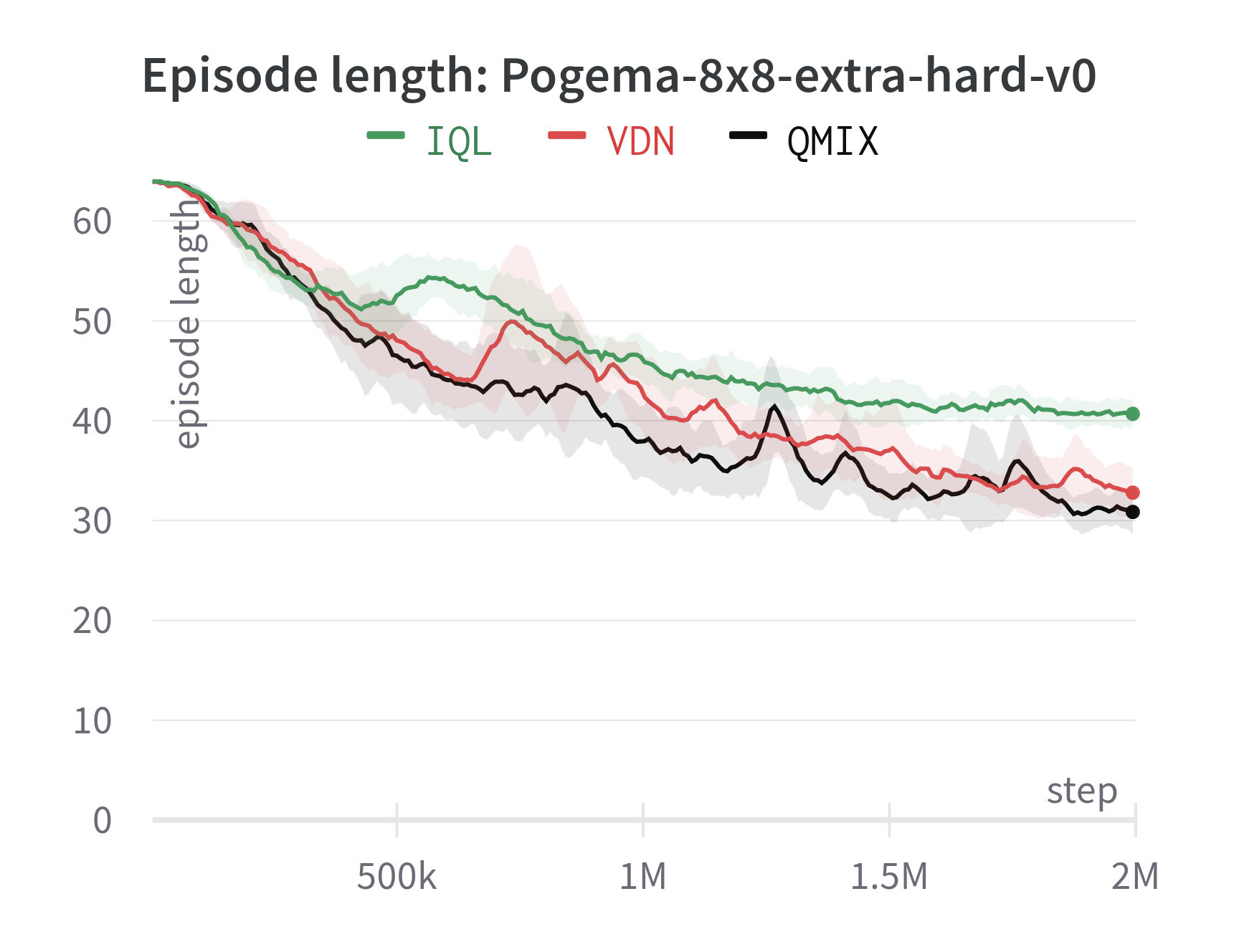}
        \end{subfigure}

        \caption{Results for QMIX, VDN and IQL for the all difficulties of $8\times8$ benchmark. There is only one agent in \textit{easy} configuration, thus we report combined plot for ISR/CSR metrics.}
        \label{fig:pymarl:all}
    \end{figure*}

    \begin{figure*}[ht!]
        \centering
        
        \begin{subfigure}[b]{0.31\textwidth}
            \includegraphics[width=\textwidth]{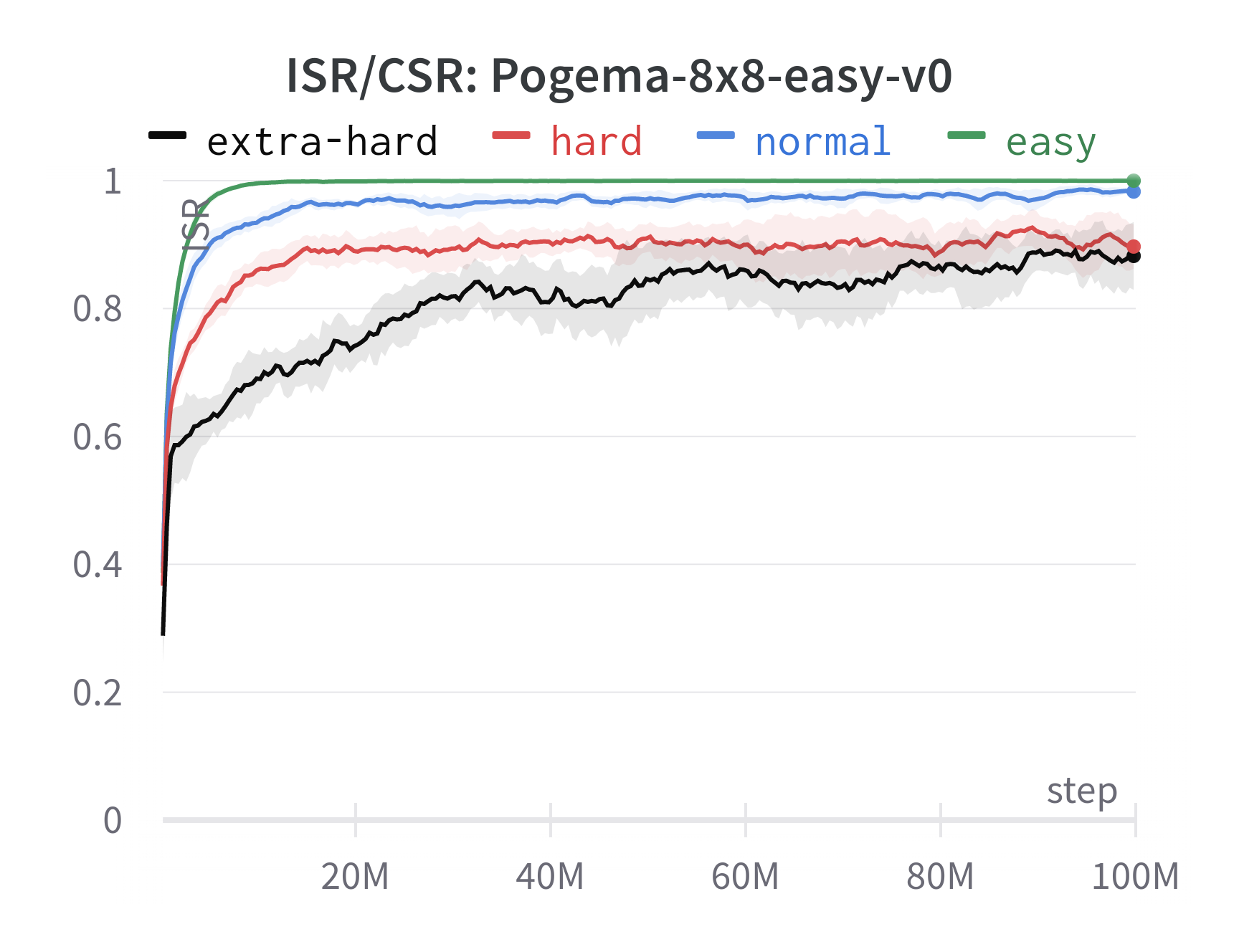}
        \end{subfigure}
        \hspace{0.01\textwidth}
        \begin{subfigure}[b]{0.31\textwidth}
            \includegraphics[width=\textwidth]{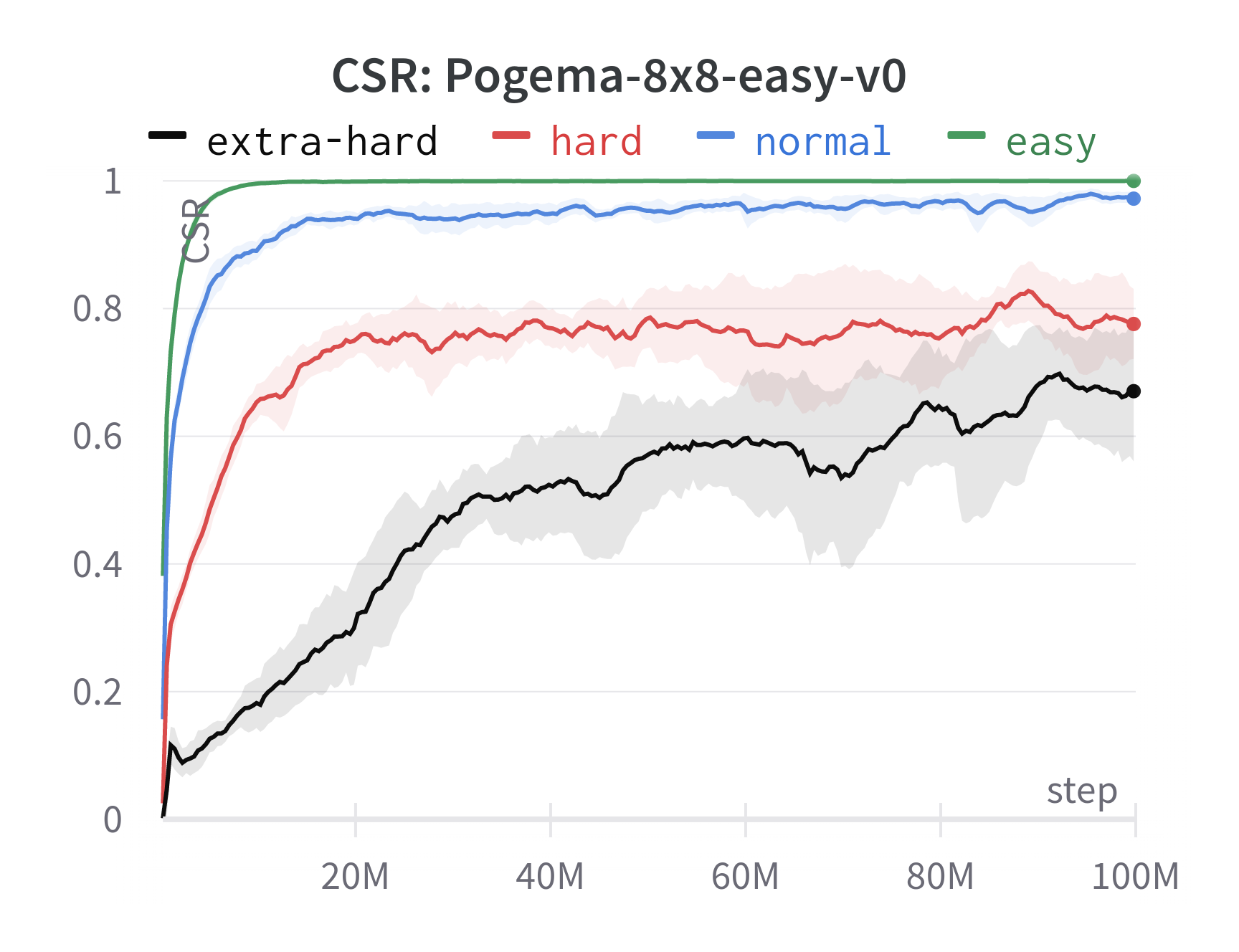}
        \end{subfigure}
        \begin{subfigure}[b]{0.31\textwidth}
            \includegraphics[width=\textwidth]{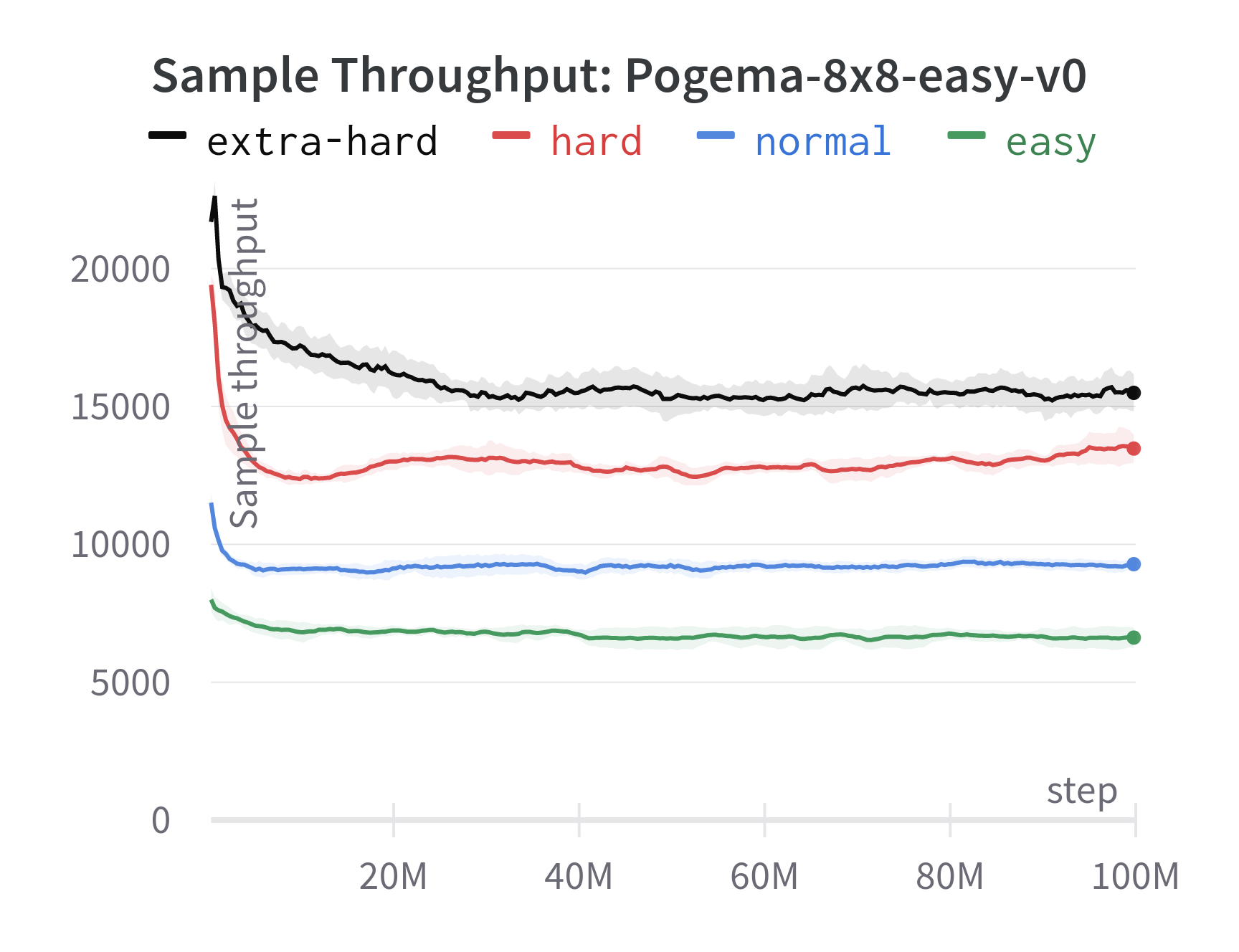}
        \end{subfigure}
        
        \begin{subfigure}[b]{0.31\textwidth}
            \includegraphics[width=\textwidth]{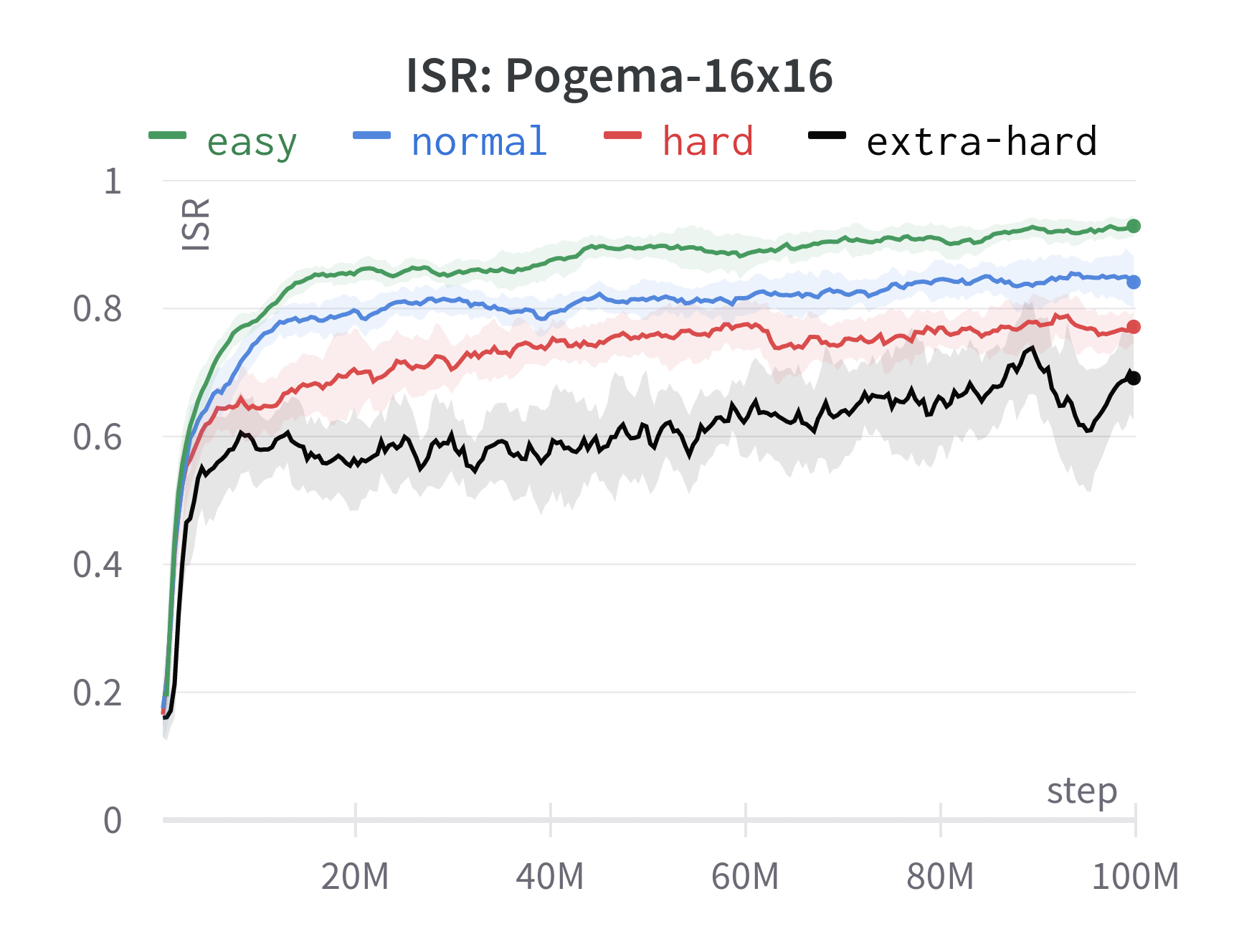}
        \end{subfigure}
        \hspace{0.01\textwidth}
        \begin{subfigure}[b]{0.31\textwidth}
            \includegraphics[width=\textwidth]{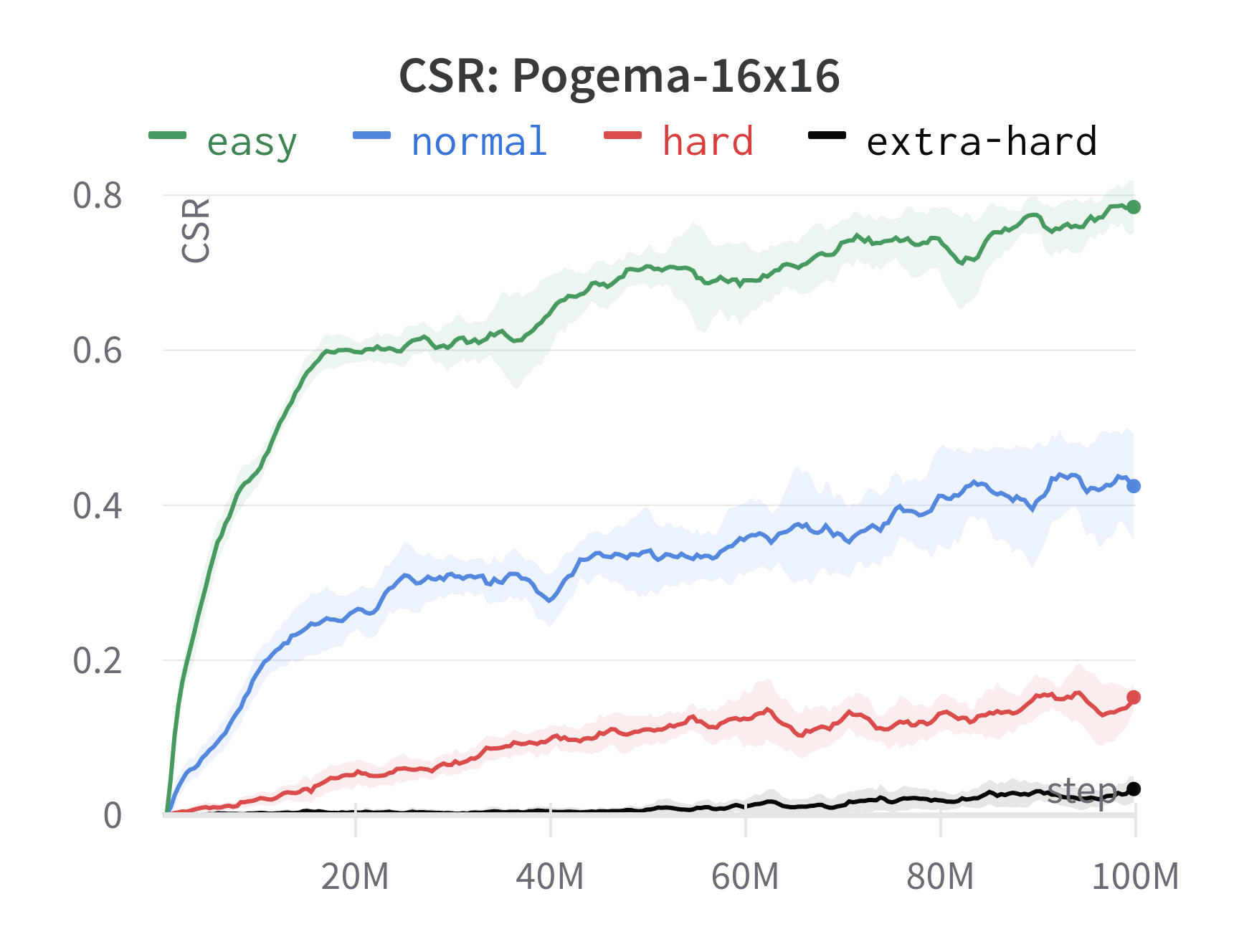}
        \end{subfigure}
        \begin{subfigure}[b]{0.31\textwidth}
            \includegraphics[width=\textwidth]{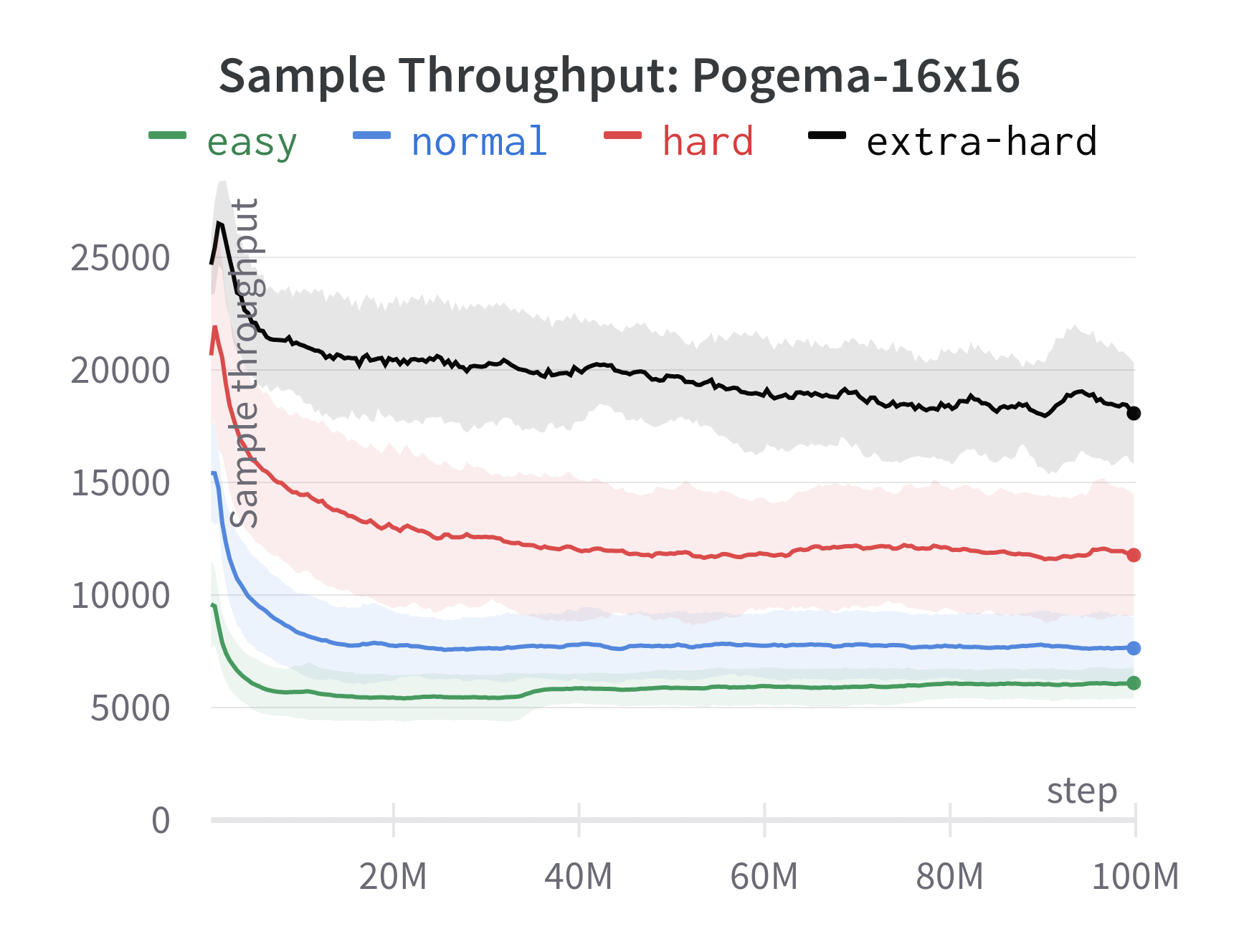}
        \end{subfigure}
        
        \begin{subfigure}[b]{0.31\textwidth}
            \includegraphics[width=\textwidth]{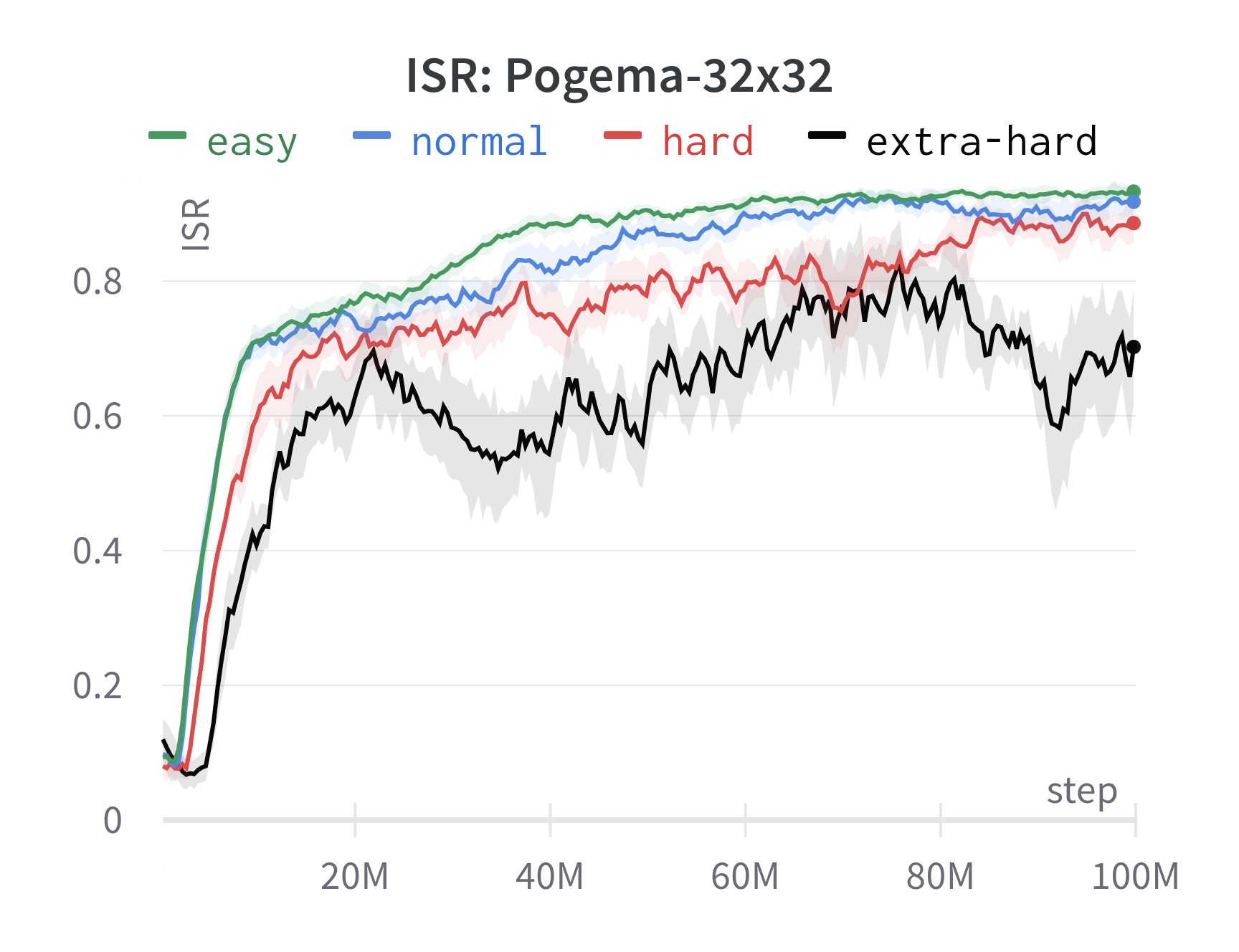}
        \end{subfigure}
        \hspace{0.01\textwidth}
        \begin{subfigure}[b]{0.31\textwidth}
            \includegraphics[width=\textwidth]{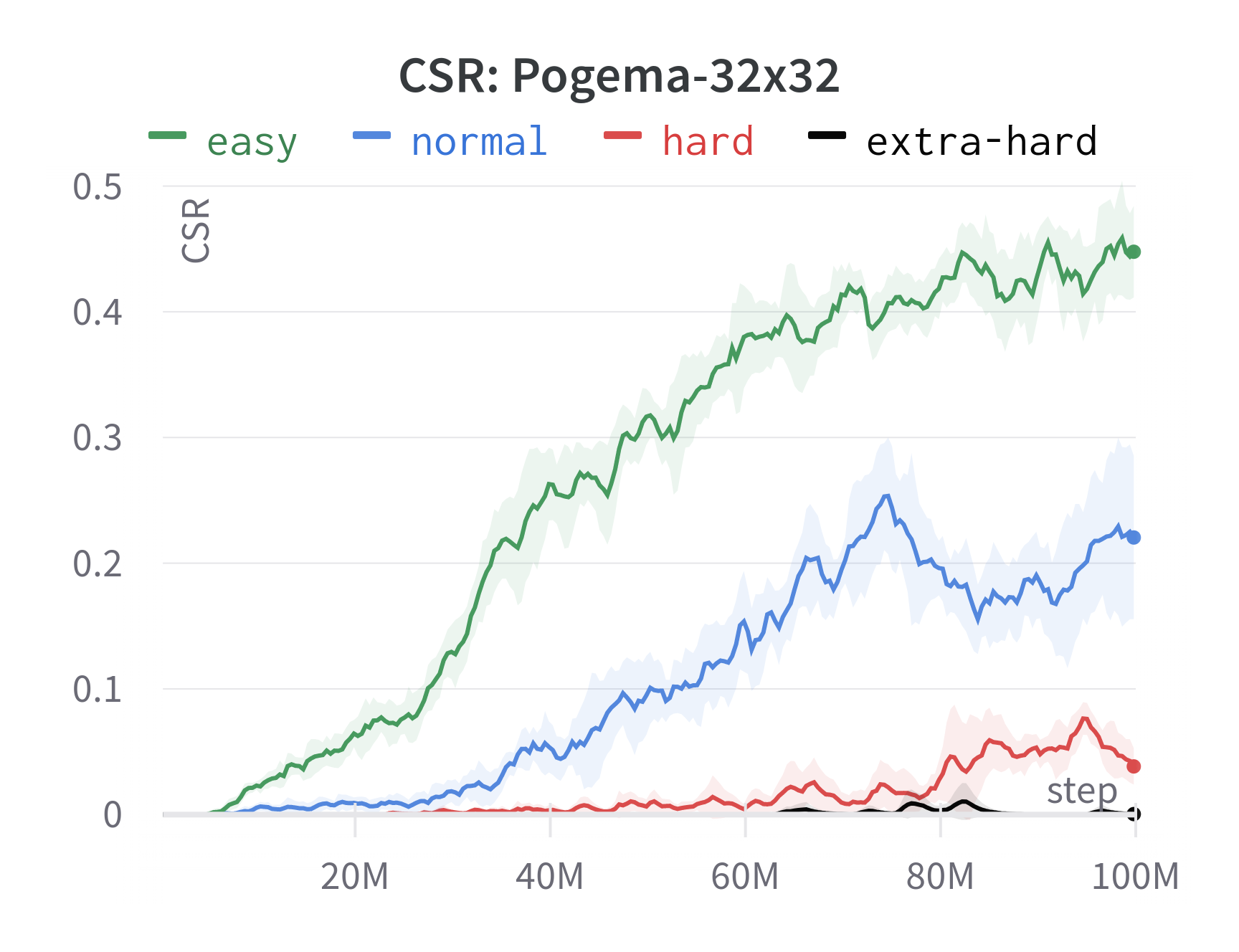}
        \end{subfigure}
        \begin{subfigure}[b]{0.31\textwidth}
            \includegraphics[width=\textwidth]{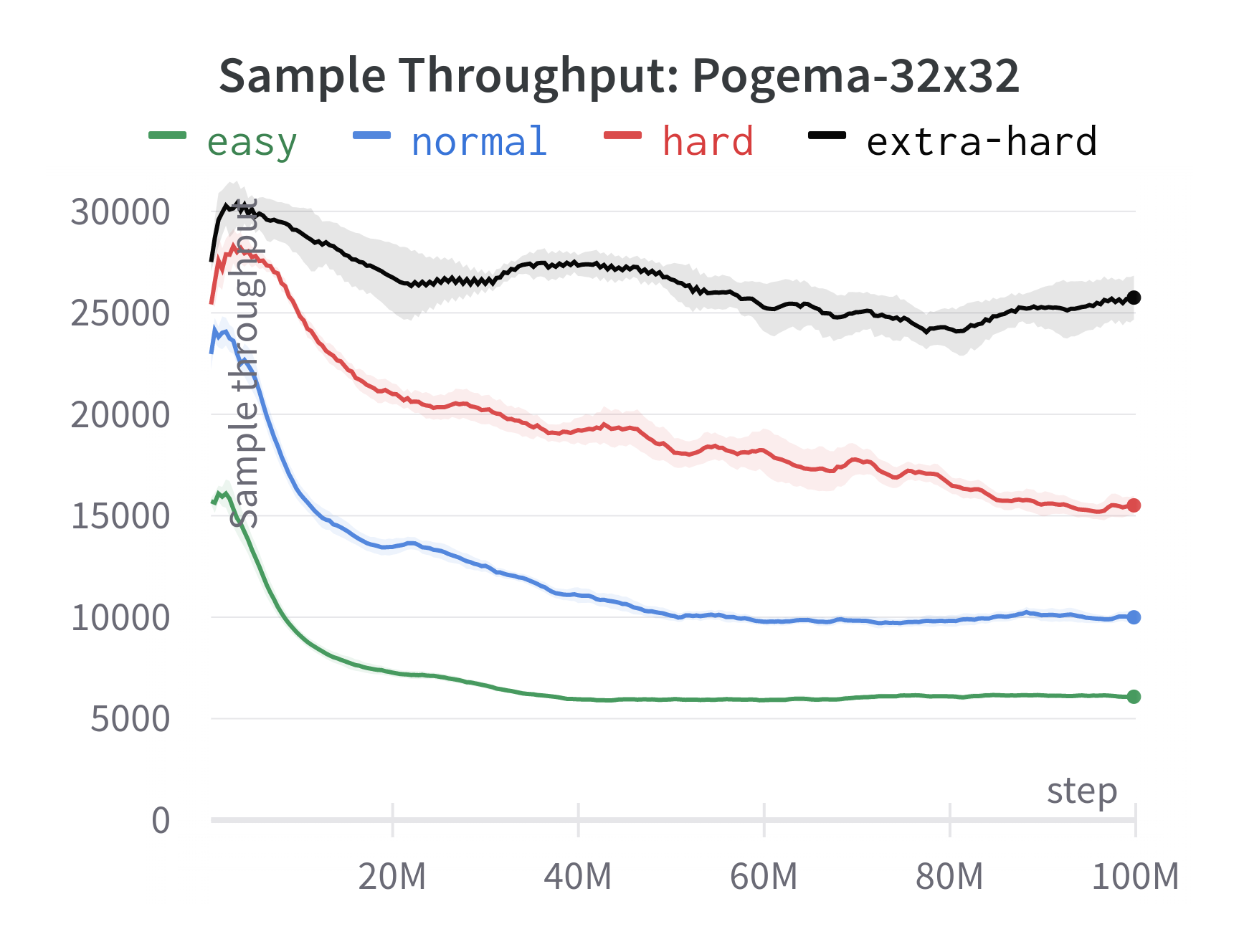}
        \end{subfigure}
        
    \caption{Results for APPO for the all difficulties of $8\times8$, $16\times16$, $32\times32$ benchmarks. Rights plots reports sample throughput for APPO.}
        \label{fig:appo:all}
    \end{figure*}
    
\end{document}